\documentclass[11pt]{article}

\usepackage[letterpaper,margin=1in]{geometry}
\usepackage[T1]{fontenc}
\usepackage{lmodern}
\usepackage{microtype}
\usepackage{amsmath,amssymb,amsthm}
\usepackage{graphicx}
\usepackage{booktabs,longtable,tabularx,array}
\usepackage{xcolor}
\usepackage[numbers,sort&compress]{natbib}
\usepackage[hidelinks]{hyperref}

\theoremstyle{plain}
\newtheorem{theorem}{Theorem}[section]

\theoremstyle{definition}

\providecommand{\Description}[2][]{}

\usepackage{booktabs}
\usepackage{longtable}
\usepackage{tabularx}
\usepackage{xcolor}
\usepackage{amsmath}

\newcommand{\FAR}{\mathrm{FAR}}

\title{Adversarial Multiplicity: Cherry-Picked Models and the Limits of Disclosure}

\newcommand{\authorrow}{\end{tabular}\par\vspace{1.5ex}\begin{tabular}[t]{c}}
\author{
  Cassandra C. Chou\thanks{Equal contribution.}\\
  Harvard University\\
  \texttt{cassiechou@g.harvard.edu}
  \and
  Benjamin Huynh\footnotemark[1]\ \thanks{Corresponding author.}\\
  Johns Hopkins University\\
  \texttt{bhuynh@jhu.edu}
  \authorrow
  Allison Koenecke\\
  Cornell Tech\\
  \texttt{koenecke@cornell.edu}
  \and
  Elizabeth T. Chin\\
  Johns Hopkins University\\
  \texttt{etchin@jhu.edu}
}

\date{}

\begin{document}

\maketitle

\begin{abstract}
Predictive multiplicity---where similarly accurate models disagree on individual predictions---can be leveraged to optimize secondary metrics (e.g., fairness) without loss in predictive performance. However, exploiting multiplicity opens the opportunity for motivated parties (or coding agents gaming their objectives) to steer predictions toward their own ends, such as in litigation or consumer disputes with automated decision-making tools. In these adversarial multiplicity settings, we distinguish soft manipulation, where parties optimize over innocuous model choices (e.g., hyperparameter tuning, feature engineering), from hard manipulation, where parties directly perturb model parameters. We find a sufficiently flexible model can be manipulated to reverse any individual prediction with minimal change in predictive performance while passing black-box audits, making hard manipulation viable in most regulatory settings. Soft manipulation has smaller effects but can pass white-box audits requiring full access to code and data. Even aggregate estimates can be manipulated: a model's estimated effect of climate change on wildfires can be fully reversed by hard edits and soft search. We find the Reconcile algorithm typically resolves most disagreements between models, but more often in favor of the manipulator when the opposing model is chosen adversarially compared to at random. We map which manipulations can plausibly pass existing regulations and disclosure policies, and recommend mitigations such as pre-registration, proof of model training, sensitivity analyses, multi-metric evaluations, and committing to versioned public models.
\end{abstract}

\section{Introduction}
\label{sec:intro}

Recent work in the algorithmic decision-making literature has focused on the issue of predictive multiplicity, where multiple models with equivalent accuracy for a given task can provide contradictory outputs \citep{marx2020,black2022,roth2023,behzad2025,ganesh2023randomness,hsu2022,ganesh2025}. As machine learning-based algorithms are increasingly used in high-stakes applications, the possibility of multiplicity raises concerns of fairness and robustness for such deployed algorithms. Numerous techniques and frameworks have been developed, such as ways to measure multiplicity \citep{marx2020, hsu2022, watsondaniels2023, hsu2024}, resolve disagreements \citep{roth2023, behzad2025, du2024}, or choose an improved model at no cost in accuracy, such as optimizing for fairness while holding predictive performance constant \citep{coston2021fairs, blackldalaw2024, dai2025}. 

We consider scenarios wherein multiplicity can be leveraged by motivated actors. Model developers and firms are typically not incentivized to optimize for fairness, but for advancing their own interests. Consider a vendor for an automated decision-making tool (ADMT) being audited for fairness; or a carbon-emitting company being sued for damages from its emissions, estimated in part by a machine learning model; or even a coding agent given simple instructions to compare models to optimize a metric: multiplicity opens the opportunity to find a model that produces a self-serving output while keeping optimal accuracy. We call this practice \emph{adversarial multiplicity}.

The extent to which adversarial multiplicity can successfully be used by motivated actors depends on their respective regulatory environments. In contexts without required disclosure, actors could simply lie: they could falsify outputs as needed, and adversarial multiplicity would not be necessary. However, in most regulatory contexts, actors are required to disclose varying levels of information on their models. Adversarial multiplicity is one potential avenue by which motivated actors could bypass regulatory audits.

Our contributions are as follows. In Section 3, we introduce two types of adversarial multiplicity: hard manipulation, where model parameters are directly perturbed, and soft manipulation, where actors search over plausible configurations of models of equivalent accuracy to pick the one that serves their needs most. In Section 4, we prove that given a sufficiently flexible model, any prediction can be hard manipulated to an arbitrary value with minimal change to predictive performance. We also empirically show this on four fairness datasets: HELOC, German, Adult, and COMPAS, where we manipulate all predictions but one to reversed decisions while remaining within 1\% of the original model's Brier score. In Section 5, we empirically show how soft manipulation can be used to reverse outputs on large fractions of cases on the same four datasets, and that requiring multi-metric evaluations reduces but does not remove the impact of soft manipulation. In Section 6, we show that both hard and soft manipulation can be applied not just to individual predictions, but also statistical estimates based on aggregated machine learning outputs. We focus on a case study where machine learning is used to estimate the effect of climate change on extreme wildfire occurrence in California---such applications are high-stakes in that (1) they inform policy for climate resilience and adaptation; and (2) they can provide the evidentiary basis for climate litigation, determining the extent to which carbon emitters are liable for damages from disasters. Specifically, we apply hard and soft manipulation to the wildfire dataset, finding that both approaches can fully reverse the reported effect of climate change. In Section 7, we test \citet{roth2023}'s Reconcile algorithm---a tool used to resolve disagreements arising from multiplicity---on adversarially manipulated predictions, finding that while it does typically resolve most disagreements, it does so in favor of the manipulator far more often than chance. In Section 8, we map out existing regulatory environments and policies to examine whether they can prevent adversarial multiplicity, and recommend practices for mitigation.

\section{Related Work}
\label{sec:related}
\paragraph{Predictive multiplicity.}
 Predictive multiplicity refers to how models of similar accuracy can produce varying outputs \citep{marx2020}. The model class defined by having similar accuracy is known as the Rashomon set, defined as a set of models that have predictive performance within a given tolerance of the best achievable performance \citep{breiman2001, fisher2019all, semenova2022}. Some common elements of model development that lead to multiplicity may be considered standard or truly arbitrary \citep{damour2022}, such as using state-of-the-art model architectures, hyperparameter optimization, dataset composition, and training randomness \citep{cooper2021hpo, ganesh2023randomness}. \citet{hsu2022} estimate the scale and capacity of predictive multiplicity, called the Rashomon capacity, by designing models using adversarial weight perturbation and identifying the models with the largest shift in predicted scores without detectable loss in performance. Other metrics have also been developed for measuring multiplicity, including discrepancy, the fraction of predictions that a competing model changes relative to the baseline model, and ambiguity, the fraction of individuals whose label is flipped by at least one high-performing model \citep{marx2020}; and viable prediction range, the minimum and maximum predicted probability for an individual \citep{watsondaniels2023}. 

\paragraph{Related forms of manipulation.}
Prior work has noted in passing that multiplicity can be exploited: \citet[\S5.3]{ganesh2025} observe that multiplicity can be used to circumvent regulatory oversight, and \citet{black2022} state that a model of equal accuracy that reverses a decision always exists in theory. Related work has also demonstrated the ability of bad actors to conduct ``backdoor attacks'' to deep neural networks using targeted weighting techniques, introducing hidden behaviors while maintaining performance \citep{zhao2019, bai2021, dumford2020, hong2022}. Further, defensible reasons for model selection, such as explainability and fairness, have been identified as ways to obscure choices made with motivated reasoning \citep{sharma2025, aivodji2019, shamsabadi2022washing, meding2024fairnesshacking, simson2024onemodel, gillis2024}. Terms like X-hacking \citep{sharma2025}, fairwashing \citep{aivodji2019, shamsabadi2022washing}, fairness hacking \citep{meding2024fairnesshacking}, and D-hacking \citep{gillis2024} have been coined to describe variations of searching for models, explanations, or evaluation choices that provide desired explanations or fairness metrics.

\paragraph{Audit manipulation and proof of training.}

\citet{yan2022active} introduce manipulation-proof auditing, where a firm may switch to any model in its class so long as it agrees with every query made by an auditor. \citet{godinot2024harder} and \citet{casper2024blackbox} argue that black-box audits are insufficient for proper regulatory oversight. Additionally, if developers know what outcomes may be plausible---especially plausible based on an auditor's prior knowledge \citep{garciabourree2025robust}---they can select a model that passes that auditing and deploy a different model. Explanation methods are similarly not robust to manipulation \citep{lemerrer2020bouncer}; \citet{anders2020fairwashing} demonstrate that any model with manipulated explanations can provide the same behavior on the data as its original model. Proof-of-training strategies can verify that the model is the result of a training run \citep{choi2023potd, jia2021pol, waiwitlikhit2024zkaudit, melissaris2026zkboost}, but only certify the possibility of this computation. Because the developer provides the dataset, model run information, and the model, there are other ways of manipulating this audit that include engineering training data \citep{luck2026cmc}. Researchers have evaluated the ability of large language models (LLMs) to detect different forms of manipulation, finding that LLMs are unreliable for this task and frequently miss sabotaged models \citep{ward2025,gan2026,libon2026}. 

\paragraph{Potential remedies.}
As a potential method of increasing transparency, multiverse analysis techniques present the results from analyses across datasets or analyses using different, theoretically-justified methodology \citep{steegen2016,simonsohn2020,silberzahn2018}. \citet{roth2023}'s Reconcile algorithm is a method of resolving disagreements between models by identifying disagreement regions, using observed outcomes to update the model that is further from the average on that region, and repeating until both models agree almost everywhere \citep{behzad2025}. \citet{kroll2017} proposes ``procedural regularity,'' which would require developers to commit their model development procedure in advance of any training or evaluation. Most widely used is pre-registration, where analysts commit to an analysis plan (including model search) prior to analysis. Such approaches are commonplace in clinical trials and in the sciences \citep{ich1998e9}, and have been proposed as a defense against X-hacking and D-hacking \citep{sharma2025, gillis2024}. 

\paragraph{Attribution science.}
Attribution studies, which determine the effect of anthropogenic climate change on extreme weather events, may be used as evidence in climate litigation \citep{philip2020, stuartsmith2021}. \citet{philip2020} illustrate that choices in event definition, dataset, and model can influence the attribution conclusion and call for standardization and transparency in the procedure. Machine learning-based attribution studies train models on present-day, observed outcomes, and predict outcomes under counterfactual climates (representing the world without climate change) \citep{brown2023,trok2024}. Previous work demonstrates that attribution studies using machine learning, especially at the individual event level, are also vulnerable to model multiplicity \citep{chou2026}.

\section{Preliminaries}
\label{sec:defs}

\begin{figure}[t]
\centering
\includegraphics[width=\textwidth]{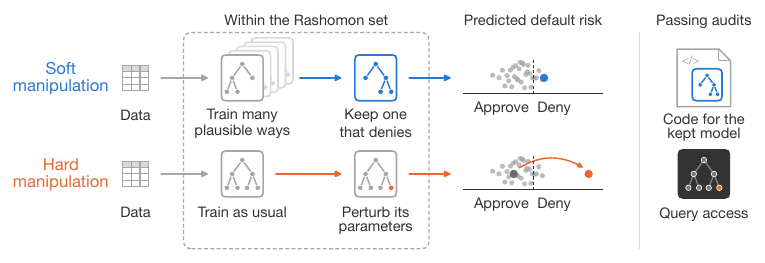}
\Description{A flow diagram in two rows, depicting soft manipulation on top and hard manipulation below. The soft manipulation row shows many trees being trained and one being kept. The hard row shows one tree trained as usual and then has its parameters edited. Each row then shows predicted default risk with an approve/deny threshold. }
\input{figs/fig1_caption}
\end{figure}

\subsection{Framework}
We define a model $f$ to be the machine learning model that maps inputs $x$ to a probability $f(x) \in [0,1]$. In binary decisions, individuals with $f(x) > t$ are considered positive cases. Models are \emph{near-equal} in predictive performance at tolerance $\varepsilon$ when their held-out performance is within $\varepsilon$ of the best-performing model, which we refer to as the \emph{reference model}. Formally, with $L$ the held-out loss and $L^*$ that of the reference model, these near-equal models comprise the Rashomon set $\mathcal{R}_\varepsilon=\{f: L(f)\le(1+\varepsilon)L^*\}$.

We define \textit{adversarial multiplicity} as the process by which a motivated actor designs or searches for a model that is within the Rashomon set and gives a desired output. We identify two approaches: \emph{hard manipulation} and \emph{soft manipulation}. Let $A$ be a training procedure (varying over model family, hyperparameters, data choices, etc.), which fits a model $A(\mathcal{D})$ on data $\mathcal{D}$. In hard manipulation, the actor reports $f_{\theta'}\in\mathcal{R}_\varepsilon$, whose parameters $\theta'$, such as tree leaf values or network weights, are edited from those of a fitted reference model $A^*(\mathcal{D})$. In soft manipulation, the actor runs many plausible procedures $A_1,\dots,A_m$ and picks $f=A_k(\mathcal{D})\in\mathcal{R}_\varepsilon$ for the $k$ that produces the desired output (Figure \ref{fig:geometry}).

\subsection{How well can audits catch manipulation?}
\label{sec:ladder-full}
Table \ref{tab:regimes} describes what types of manipulation auditors can identify depending on the amount of information model developers are required to disclose. A key assumption we make is that developers are incentivized to not get caught lying when disclosing documentation. Outright fabricating information or producing falsified model outputs may run the risk of incurring severe penalties. By contrast, hard and soft manipulation can be described in vague or broad terms such that documenting it as such (e.g., as ``post-processing'' or ``model search'') may be less likely to incur penalties. Hard manipulation would not be caught if auditors are only given query access through black box testing, as varying inputs and observing outputs would not reveal edited parameters. However, hard manipulation cannot pass through auditing that includes both training data and code, as the code would reveal manual changes to the model parameters. 

Soft manipulation can be hidden at most levels of auditing. Because the manipulation occurs at model selection and there is no direct change of model parameters, even a refit of the model using the same code and data would not reveal soft manipulation. Motivated actors could just present the model information from the chosen model without fully disclosing the model search process, avoiding auditor detection. However, pre-registration in combination with code and data audits could be used to confirm whether a disclosed model comes from the pre-registered model search plan.

\begin{table}[t]
\caption{\textbf{Which levels of model manipulation are identified based on levels of auditor access.} ``Passes'' indicates that the auditor is not able to catch manipulation, and ``identified'' indicates that the auditor is able to determine there was manipulation.}
\label{tab:regimes}
\footnotesize
\setlength{\tabcolsep}{4pt}
\begin{tabularx}{\textwidth}{@{}>{\raggedright\arraybackslash}p{0.2\textwidth}
  >{\raggedright\arraybackslash}p{0.08\textwidth}>{\raggedright\arraybackslash}X
  >{\raggedright\arraybackslash}p{0.08\textwidth}>{\raggedright\arraybackslash}X@{}}
\toprule
 & \multicolumn{2}{l}{\textbf{Hard manipulation}} & \multicolumn{2}{l}{\textbf{Soft manipulation}} \\
\cmidrule(lr){2-3}\cmidrule(l){4-5}
\textbf{The auditor receives} $\downarrow$ & \textit{Result} & \textit{Why} & \textit{Result} & \textit{Why} \\
\midrule
Documentation and outputs & Passes & Documentation can describe the edit in broad or vague terms & Passes & Model search can be broadly justified \\
\addlinespace
\hspace*{0.6em}+\,Black box testing & Passes & Queries do not reveal edited parameters & Passes & The chosen model answers queries consistently \\
\addlinespace
\hspace*{1.2em}+\,Training data and code & Identified & Code for the original, unedited model is discovered & Passes & The chosen model fits identically from the disclosed code and data \\
\addlinespace
\hspace*{1.8em}+\,Pre-registration & Identified & As above & Identified & The disclosed model must come from a pre-registered model search plan \\
\bottomrule
\end{tabularx}
\end{table}

\section{Hard Manipulation}
\subsection{Theoretical framework}

We find that via hard manipulation, any individual classification can be reversed while holding every other scored prediction fixed and staying within the Rashomon set. The conditions required are (1) the model has to be \emph{sufficiently flexible}, where, roughly, there are at least as many editable parameters as observations; and (2) manipulating a single prediction (along with any other identical observations) still keeps the model within the Rashomon set. The formal justification is as follows.

\begin{theorem}[Any prediction can be manipulated while holding all other scored predictions fixed]
\label{thm:hard}
Let a model's output be a linear function of its learned representation $\phi(x)$, with parameters $\theta$, passed through a strictly increasing function $\psi$: $f(x)=\psi(\theta^\top\phi(x))$. If, and only if, the scored data points' distinct
vectors $\phi(x)$ span as many dimensions as there are such vectors, any scored data point's prediction can be changed to any value $\psi$ takes by
changing $\theta$ alone, with every other scored prediction unchanged except those sharing its $\phi(x)$ exactly. 
\end{theorem}

The conditions for this theorem are mild. Most flexible machine learning models, such as tree-based models or neural networks, fit the condition of their final outputs being a linear function of its learned representation $\phi(x)$, such as the leaves of a tree or the last hidden layer of a neural network. In practice, meeting the conditions of ``$\phi(x)$ spanning as many dimensions as there are such vectors'' simply means having a large enough model: enough leaves in a tree-based model, or enough units in the last hidden layer of a neural network. Formally, we refer to this as the \emph{rank condition}: if one were to represent the distinct vectors $\phi(x)$ as the rows of a matrix $\Phi$, then those rows of $\Phi$ must be linearly independent---in other words, $\Phi$ must have full row rank.

It follows from this theorem that a reference model manipulated via this method will remain in the Rashomon set so long as the loss rises by no more than $\varepsilon L^*$. Typically, a targeted observation $i$ is the only row with $\phi(x_{i})$---unless there are other functionally identical rows in the dataset---so the performance loss from changing one observation is minimal. Furthermore, the scoring assumptions here are conservative: in applications, performance is typically reported on a held-out set, and if a targeted observation is not within that held-out set, then it would be trivial to manipulate the observation without any change to the reported predictive performance. The proof for Theorem \ref{thm:hard} is in Appendix~\ref{app:hard}.

\begin{figure}[t]
\centering
\includegraphics[width=\textwidth]{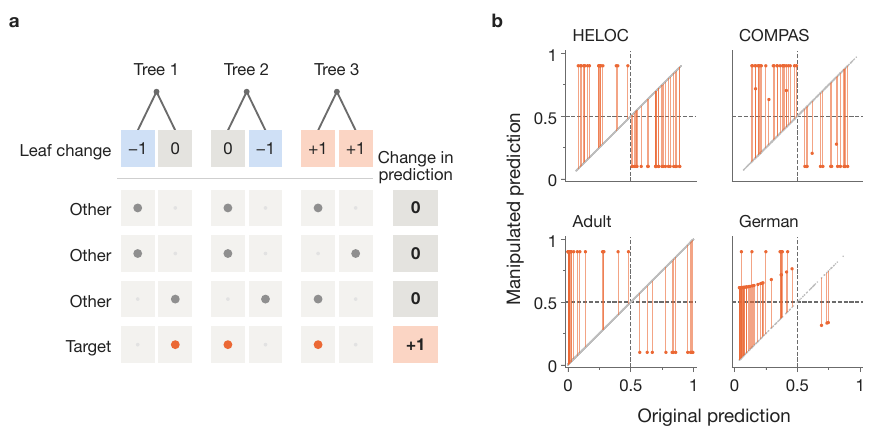}
\Description{Panel a: a grid with three trees as columns and four individuals as rows; a diagram explaining how hard manipulation works. Panel b: four scatter plots (HELOC, COMPAS, Adult, German) of manipulated against original prediction, with a grey identity line and dashed lines at 0.5.}
\caption{\textbf{Hard manipulation can change the decisions of any individual without detectable loss in performance.} \textbf{a,} Toy model schematic of how hard manipulation works. Each row represents an individual, and its prediction is the sum of the leaves the individual lands in. The changes to the leaves increase the target's prediction by +1 while canceling across the other individuals. \textbf{b,} Hard manipulation on four fair machine learning datasets. Grey line indicates the original predictions; if no predictions were changed after manipulation, we would only observe this $y = x$. Each orange point represents the changed prediction from a target individual in the validation dataset---all individuals had reversed decisions after manipulation. All models with changes were kept within the Rashomon set.}
\label{fig:hard}
\end{figure}

\subsection{Application of hard manipulation}
\label{sec:hard-app}
We demonstrate that hard manipulation of any individual prediction is possible using four commonly used datasets in fair machine learning: HELOC \citep{davis2022explainable}, COMPAS \citep{larson2016we}, Adult \citep{ronny1996uci}, and Statlog (German credit data) \citep{statlog_german_credit_data_144}. HELOC predicts whether credit applicants to FICO's home-equity line will repay using data from roughly 10,000 applications. COMPAS predicts whether defendants will be re-arrested within two years using data from 5,278 defendants from Broward County, Florida. Adult predicts whether an individual's income is above \$50,000 using data from the 1994 United States Census---we use 48,842 individuals. Statlog's German credit data predicts whether an individual has good or bad credit using 1,000 loan applicants from a German bank. Each dataset is split into training, validation, and test rows using a 60/20/20 split. For this analysis, we only assess hard manipulation on the scored validation rows, as doing so on the test rows would be trivial (one could manipulate any observation outside of the test set without any loss in held-out performance). 

The reference model is a LightGBM \citep{ke2017lightgbm} model trained by a standard tuning search using 5-fold cross-validation with log loss. To identify our starting model for manipulation, we train several LightGBM models on the training split of the data and determine the Rashomon set---the set of candidate models whose validation Brier score (the mean squared difference between the predicted probability and the binary 0/1 outcome; lower is better) is within $\varepsilon = 1\%$ of the reference model's validation Brier.\footnote{The choice of model and loss function for this experiment is arbitrary and for illustrative reasons only. In practice, one could do a full search across all model families and then apply hard manipulation to whichever model has the lowest loss.} Notably, the Brier score may increase or decrease; the original prediction may get flipped towards the true label if it was on the wrong side of the decision threshold originally, leading to a decrease in Brier score (and vice versa). Then, we select a model within this range that meets the rank condition. We define a group to be a set of training and validation rows that land in the same leaf in every tree, meaning they always change together; groups that do not meet the rank condition cannot be manipulated without shifting the predictions of other individuals, and therefore cannot be changed.

For each individual in the validation set, we find a perturbation of the leaf values that shifts the prediction of the target individual while leaving all other individuals in the training and validation set unchanged (assuming the individual does not belong to a group). Moving the prediction of the target individual changes the Brier score by design, so we find the largest possible change to the individual that flips the individual's prediction across the decision threshold of 0.5 for all datasets---up to 0.9 (or 0.1)---that does not take our model outside of the Rashomon set.\footnote{The manipulations to 0.9 and 0.1 are also arbitrary---here we just aim to show that predictions can be manipulated to extremely high or low values.} The magnitude of the individual's change in prediction is therefore fixed, and we chose to implement the perturbation with the smallest manipulations in Euclidean norm measured over all leaves together. For 40 randomly chosen target individuals in the validation set for each dataset, we apply this manipulation to the model to empirically confirm the hard manipulation (Figure \ref{fig:hard}). 

\subsection{Individual predictions can be manipulated without detectable loss in performance}
\label{sec:hard}
We find that, with the exception of one individual who does not meet the rank condition in the Adult dataset, \emph{the decision of any validation individual across datasets can be reversed with minimal change in performance using hard manipulation} (Figure \ref{fig:hard}b). In terms of whether those predictions can be manipulated to arbitrarily high or low values well beyond the decision threshold (altering the prediction to 0.9 or 0.1), we find that it can be done for 99.9\% of all validation individuals in the HELOC and Adult datasets, 91\% in the COMPAS dataset, and 30\% in the German dataset. This is due to the fact that there are only 200 individuals in the German validation set, so each individual is responsible for a greater proportion of the Brier score. In practice, individuals in the test set may be flipped as well; this is a more likely scenario as validation performance metrics do not necessarily change with manipulation of test set individuals. This is also the more trivial case, as the target's change in prediction would not affect the validation Brier score. 

\section{Soft Manipulation}
\label{sec:soft}
Using the same four datasets (HELOC, COMPAS, Adult, and German), we evaluate the ability of soft manipulation to reverse the decision of individual predictions. For each of the four datasets, we simulate the process of a model developer by designing candidate models. We make data processing choices (such as pruning correlated features, dropping missing values), fit 6 different families of models (including LightGBM \citep{ke2017lightgbm}, XGBoost \citep{chen2016xgboost}, CatBoost \citep{prokhorenkova2018catboost}, random forests \citep{breiman2001random}, multi-layer perceptrons \citep{rumelhart1986learning, pedregosa2011scikit}, and TabPFN \citep{hollmann2022tabpfn}\footnote{We chose these models to represent the state-of-the-art (tree-based models, neural networks, and a tabular foundation model) widely used for tabular prediction.}), vary across hyperparameters, use different seeds for training, and use four methods of calibration (none, Platt \citep{platt1999probabilistic}, isotonic \citep{niculescu2005obtaining}, spline \citep{lucena2018spline}), with each fitted on a held-out 20\% fold of training rows. To identify our reference model, which reflects the model chosen by a good-faith analyst with no stake in any particular individual, we use the model with the lowest validation Brier score from 30 random configurations of data choices, hyperparameters and seed for each of five families, each under the four calibrations for a total of 600 candidate models, plus every setting of TabPFN's grid fitted once (16 fits on German and COMPAS, 144 on HELOC, 32 on Adult). We create a larger pool of candidate models to replicate adversarial model selection: 450 configurations (instead of 30) per family for the five families, every TabPFN setting with three seeds, and 20 seed-only refits of each family's best standard-search model, for a total of 2,398 to 2,782 fits per dataset, each scored under the four calibrations.\footnote{TabPFN did not include calibration methods due to computational cost; initial runs indicated calibration did not improve performance for TabPFN.}

\begin{figure}[t]
\centering
\includegraphics[width=\textwidth]{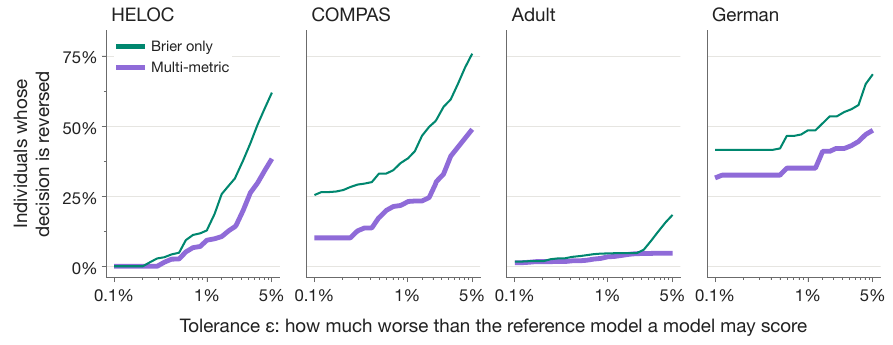}
\Description{Four line plots, one per dataset (HELOC, COMPAS, Adult, German). The x-axis represents tolerance on a log scale from 0.1\% to 5\%. The y-axis is the percentage of individuals whose decision is reversed. A green line shows the Brier-score-only check and a purple line shows the multi-metric check.}
\caption{\textbf{Soft manipulation may reverse the decision of some individual predictions, even under multi-metric evaluation.} Reach, or percentage of target individuals whose decision is reversed, for different levels of tolerance (how much worse a chosen model can score in comparison to the reference model) for four fair machine learning datasets. Includes soft manipulation under a Brier score only check and a multi-metric check.}
\label{fig:soft}
\end{figure}

\subsection{Adversarial model selection can reverse the decision of individual predictions}
We define \textit{reach} as the percentage of individuals in the test set for whom there exists at least one model in the Rashomon set of the actor's large pool of candidate models that gives the opposite decision to the reference model. The Rashomon set includes any model with a validation Brier score within a preset $\varepsilon$ of the reference model---we use a range of 26 $\varepsilon$ values from 0.1\% to 5\%. We use a single metric, the Brier score, because single-metric model selection is common in practice and is how the Rashomon set is typically defined \citep{semenova2022, marx2020}. We explore multi-metric evaluation in the following section (Section~\ref{sec:multi-metric}).

Our results demonstrate that soft manipulation of test individuals is possible at several different epsilon levels. At $\varepsilon = 5\%$, reach is 61.9\% of 1,899 HELOC applicants (1,972 models in Rashomon set), 75.8\% of 563 COMPAS defendants (1,405), 18.3\% of 9,761 Adult individuals (700), and 68.5\% of 200 German credit applicants (294). We find that at $\varepsilon = 1\%$, reach is 12.8\% for HELOC (56), 38.4\% for COMPAS (275), 4.6\% for Adult (90), and 48.5\% for German credit (26). Even at $\varepsilon = 0.1\%$, reach is 25.4\% for COMPAS (30) and 41.5\% for German credit (12), against 1.7\% for Adult (8) and 0\% for HELOC (0).
 
\subsection{Using multiple performance metrics for evaluation reduces soft manipulation, but does not remove it}
\label{sec:multi-metric}
We assess whether adding additional validation performance metrics as criteria for being in the Rashomon set reduces the effect of soft manipulation. We add four checks (in addition to Brier score): $1-\text{AUC}$ within $\varepsilon$ of the best $1-\text{AUC}$ on the validation set, validation calibration-in-the-large such that $|\frac{\text{Mean Predicted Probability}}{\text{Mean Observed Outcome}} - 1|$ is within 0.1 of the best calibration-in-the-large, calibration slope such that $|\text{slope} - 1|$ is within $0.1$ of the best calibration slope, and validation impact ratio, the favorable-decision rate of the least-favored group divided by that of the most-favored group for a protected attribute, which must not be significantly worse than the reference model's (see Appendix~\ref{supp:multi-metric}). These metrics were chosen to represent additional standard performance metrics (AUC, calibration-in-the-large, calibration slope) and follow existing guidance and regulations for demonstrating a model is nondiscriminatory (impact ratio). Specifically, the impact ratio is the quantity that algorithmic audits of hiring tools must report under New York City's Local Law 144 \citep{nycdcwp2023aedt}, as well as the basis of the \emph{four-fifths rule} for adverse impact in the federal Uniform Guidelines on Employee Selection Procedures \citep{uniformguidelines1978}.

Requiring models to pass the multi-metric check lowers reach on all datasets, but does not remove the ability of soft manipulation to flip individual decisions. At $\varepsilon = 1\%$, reach falls from 12.8\% to 9.3\% on HELOC, 38.4\% to 22.9\% on COMPAS, 4.6\% to 3.4\% on Adult, and 48.5\% to 35.0\% on German credit. This pattern holds across $\varepsilon$ values (Figure~\ref{fig:soft}). The effect of each metric varies by dataset; without AUC at $\varepsilon = 1\%$, reach increases by 3.4pp on HELOC, 9.8pp on COMPAS, 1.2pp on Adult, and 4.5pp on German. Additionally, without the calibration check, reach on German increases by 9.5pp but has near-negligible effects (0 on HELOC and Adult, +0.2pp on COMPAS) on the other datasets. Regardless, multi-metric evaluation mitigates some of the effect of soft manipulation, but does not remove it completely.

\section{Manipulation of Aggregate Estimates: Wildfire Attribution}
Aggregate estimates, which combine predictions from multiple individuals into a single estimate, are also vulnerable to manipulation. We assess this using California wildfire data from 2003--2020, estimating the effect of anthropogenic climate change on wildfire growth (represented as a function of the ratio between observed present-day outcomes and counterfactual pre-industrial outcomes) using an approach similar to \citet{brown2023}. We define wildfire growth as growth over 10,000 acres in a single day, and our observed dataset includes 17,910 fire days (380 extreme growth days). 

We fit models to the observed, present-day data and use the models to predict outcomes in a pre-industrial counterfactual climate scenario---a climate scenario we never observe. Each day in the pre-industrial climate scenario has a corresponding day in the present-day scenario; all the covariates (moisture, precipitation, wind speed, land use category (forest, shrub, or savanna/grassland), land aspect (cardinal direction), land slope, elevation above sea level, and vegetation fraction) are matching except for the climate variables (temperature, dead fuel moisture, and vapor pressure deficit). Temperature is changed using general circulation models and vapor pressure deficit and 100- and 1000-hour dead fuel moisture are transformed from the new temperature using physics-based equations to represent a pre-industrial time period \citep{brown2023}. To avoid training and validating on matching days, we follow the cross-validation scheme in \citet{chou2026} and fit our model six times using a temporal (split by year) leave-one-block-out cross-validation. We use the model trained on the five other time blocks in the observed dataset to predict on the last block in both the present-day and pre-industrial climate scenarios. 

Formally, for each fire day $i$, let $x_{i0}$ be its observed present-day features, $y_{i0} \in \{0,1\}$ a binary indicator denoting whether extreme growth occurred, and $x_{i1}$ the counterfactual features under a pre-industrial climate. We fit a model $\hat f$ on the observed pairs $(x_{i0}, y_{i0})$ and apply it to the counterfactual features to get $\hat p_{i1} = \hat f(x_{i1})$. 
Our aggregate estimate of interest is the fraction of attributable risk (FAR), which can be represented as $\FAR = 1 - \bar{p_1}/\bar{p_0}$ where $\bar{p_1}$ is the average predicted probability of wildfire growth in the pre-industrial climate scenario ($\bar{p_1} = \frac{1}{n}\sum_{i= 1}^{n}\hat{p}_{i1}$), and $\bar{p_0}$ is the observed probability of wildfire growth in the present-day climate scenario ($\bar{p_0} = 380/17910 = 0.0212$) \citep{philip2020}.

\begin{figure}[t]
\centering
\includegraphics[width=\textwidth]{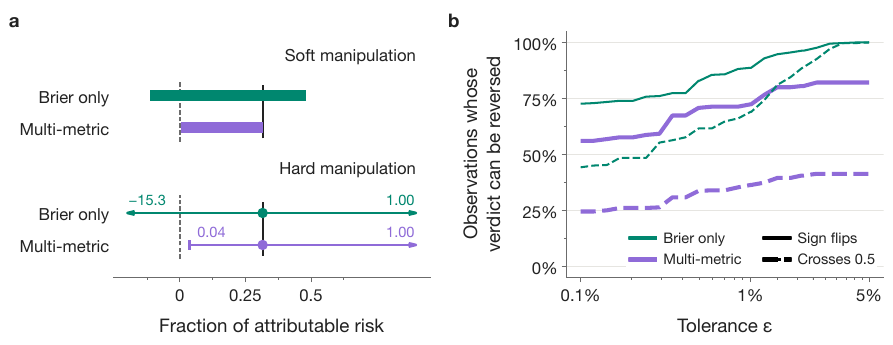}
\Description{Panel a: horizontal interval bars of the fraction of attributable risk, for soft manipulation (Brier only and multi-metric) and hard manipulation (Brier only and multi-metric), with a dashed line at zero and a solid line at the reference estimate. Panel b: lines of the percentage of extreme fire days whose attribution can be reversed against tolerance on a log scale, solid for sign flips and dashed for crossing 0.5, in green for Brier only and purple for multi-metric.}
\caption{\textbf{Attribution of extreme wildfire growth to climate change under soft and hard manipulation.} \textbf{a,} Range of aggregate fraction of attributable risk (FAR, pre-industrial vs. present-day) estimates at $\varepsilon = 1\%$ under soft and hard manipulation using the Brier score-only and multi-metric evaluation. Reference model's FAR is 0.32. Hard manipulation values are clipped on the negative x-axis. \textbf{b,} Percentage of the extreme fire days whose decision can be reversed (FAR crosses 0) under soft manipulation using Brier score-only and multi-metric evaluation. Additionally includes percentage of extreme fire days where the FAR may cross 0.5, the threshold often used to denote legal responsibility.}
\label{fig:wf}
\end{figure}

\subsection{Hard manipulation still reverses aggregate estimates}
\label{sec:wildfire-hard}
We apply hard manipulation in a similar method to Section \ref{sec:hard-app}. Using the reference model from the soft manipulation application described below in Section \ref{sec:wildfire-soft}, which is a CatBoost model, we fit one model per temporal fold. The predicted probability of extreme fire growth on a fire day is an out-of-sample prediction from its fold's corresponding model, equivalent to a logistic function of the leaf values from each tree in the model. The manipulation is applied to the leaf values, leaving each tree fixed. We find a set of changes on each fold such that the cross-validated Brier score over all 17,910 present-day days (all folds combined) stays within the 1\% bar while moving the mean pre-industrial prediction up or down. We apply one set of leaf-value changes to all six fold models jointly, and re-predict the probability of extreme fire growth in the present-day and pre-industrial climate scenarios (see Appendix~\ref{supp:wildfire-hard}).

We find that the aggregate FAR estimates can be manipulated: the original FAR was 0.32 and the lowest manipulated FAR was $-15.3$ (reached $-15.3$ to 1.0), reversing the implication of the effect of climate change on wildfire growth. However, when using the multi-metric check to identify the Rashomon set (see Section~\ref{sec:multi-metric}), the FAR cannot be manipulated to change directions: the range of possible predictions is 0.04 or up to 1.00 (Figure \ref{fig:wf}a). In practice, FAR of $-15.3$ has an extremely unrealistic interpretation; it would indicate that climate change has made the likelihood of extreme wildfire growth 16 times less likely. However, this application illustrates that hard manipulation can be used to achieve effectively any desired aggregate estimate, which could potentially be misused by actors incentivized to downplay the effects of climate change. Multi-metric evaluation can mitigate the effect of such manipulations, but not remove it entirely.

\subsection{Soft manipulation alters aggregate estimates}
\label{sec:wildfire-soft}
We follow the same methodology in Section \ref{sec:soft} with the additional temporal fold structure to apply soft manipulation to the wildfire dataset. Five model families are searched (TabPFN was excluded due to computational cost). The reference model is chosen from the set of models using 30 configurations per model family (150 fits, 600 model--calibration candidates) using the best Brier score across the whole present-day dataset. The adversarial model is chosen from the set of models using 150 configurations per model family (150 configurations per family plus 10 seed-only refits of each family's best standard-search model; 3,200 candidate models) that fall within $\varepsilon = 1\%$ of the reference model. The adversarial model is the model within this Rashomon set with the lowest aggregate FAR.

In the aggregate FAR case, soft manipulation can reverse the implication for climate change: within the Rashomon set, the FAR ranges from $-0.11$ to 0.48 around the reference model's 0.32. When using a multi-metric evaluation (see Section \ref{sec:multi-metric}), FAR still ranges from 0.01 to 0.32. As before, the focus is not necessarily on whether the FAR can be flipped directionally (as it's physically implausible that the FAR would be negative), but rather that the estimated FAR can be manipulated to give nearly any arbitrary value up to zero. In the context of climate litigation, an adversarially manipulated reduction in estimated FAR could lead to lower estimated damages and therefore less accountability for a carbon-emitting defendant.

\subsection{Soft manipulation can reverse decisions on individual fire days}
We can also use soft manipulation on individual fire days; instead of choosing models from the Rashomon set using the aggregate FAR, we choose the model for each extreme fire day that shifts the day's FAR ($\FAR_i = 1 -\hat{p_{i1}}/\hat{p_{i0}}$) the furthest. We find that on 89\% of the 380 extreme fire days ($\varepsilon = 1\%$), there exists a model that reverses the sign of the FAR. Using multi-metric evaluation, this model exists for 72\% of extreme fire days (Figure \ref{fig:wf}b). Additionally, we find that on 69\% of extreme fire days, using the Brier-only evaluation, there exists a model that gives an FAR value greater than 0.5. An FAR estimate of greater than 0.5 indicates that climate change is ``more likely than not'' responsible for an event occurring, which is typically the bar at which climate change is legally considered responsible for extreme events \citep{lloyd2021climate, lloyd2021proof, burger2019law}. Using the multi-metric evaluation, 36\% of extreme growth days have a model within the Rashomon set with FAR $> 0.5$. This is especially significant for climate litigation, where cases usually focus on a single extreme event (which would correspond to a single extreme fire day). Again, we find here that multi-metric evaluation can help mitigate the effects of soft manipulation, but cannot remove its reach completely.

\section{Reconcile Against Adversarial Multiplicity}
\label{sec:reconcile}
We assess the ability of the Reconcile algorithm to combat adversarial multiplicity using the four datasets (HELOC, COMPAS, Adult, and German). We follow a similar method to Section \ref{sec:soft} to choose a reference model. Instead of using the full validation set to calculate the Brier score, we split the validation set into two sets (A and B) and use half of the validation set (A). This prevents leakage from the model selection into the Reconcile method, as without the split, Reconcile would favor the reference model's predictions by construction. The reference model is the model with the lowest Brier score on validation set A, and the Rashomon set is the set of models with Brier scores on validation set A within $\varepsilon = 1\%$ of the reference model. We repeat this process for 10 random splits, choosing the reference model and the Rashomon set in each, and report each outcome averaged over the 10 splits. Appendix~\ref{supp:reconcile} provides further info.

For each individual in the test set, the outcome is one of three options: changed, where both reconciled models give the reversed decision; original, where both reconciled models give the reference model's decision; or disputed, where the two reconciled models disagree. We use two methods of selecting partner models: adversarial and arbitrary. The adversarial model is the model that, for each individual in the test set, reverses the decision and has the largest magnitude in prediction (largest $|\hat{p}-0.5|$) before running the Reconcile method. For the arbitrary partner, we run Reconcile on the reference model paired with every model in the Rashomon set and calculate the percentage of outcomes in each category (changed, disputed, original) for each individual. We then take the mean percentage in each category across all individuals in the test set, representing the exact expected outcome for a partner drawn at random from the Rashomon set (whether or not the partner reverses the individual's decision).

\begin{figure}[t]
\centering
\includegraphics[width=0.72\textwidth]{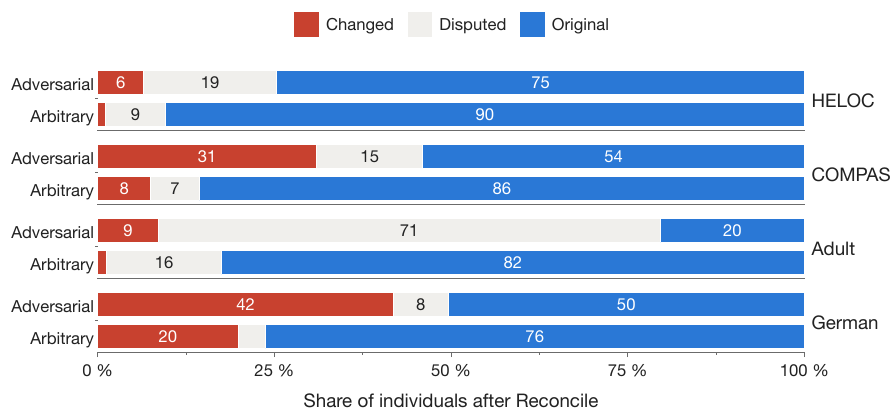}
\Description{Stacked horizontal bar chart with two bars per dataset (HELOC, COMPAS, Adult, German): one for an adversarially chosen partner model and one for an arbitrarily chosen partner.}
\caption{\textbf{Adversarial model decisions survive reconciliation more often than arbitrary model decisions.} Percentage of individuals that fall into each category after Reconcile, when the partner is chosen adversarially or arbitrarily. Changed indicates that both models give the partner model's decision; disputed indicates both models retain their original, split decisions; original indicates that both models give the reference model's decision.}
\label{fig:rec}
\end{figure}

When the adversarial model is chosen as the reference model's partner for Reconcile, the adversarial model's original decision survives---defined as giving its original adversarial decision after reconciliation---more frequently than an arbitrary partner. The adversarial decision flips the reference model's decision 6.1 times (6.5\% vs. 1.07\%, adversarial vs. arbitrary), 4.1 times (30.99\% vs. 7.52\%), 6.7 times (8.55\% vs. 1.27\%), and 2.1 times (41.85\% vs. 19.88\%) the arbitrary decision after Reconcile on HELOC, COMPAS, Adult, and German respectively. Further, the adversarial model's decision remains disputed (with the two reconciled models still disagreeing) 2.2 times (18.8\% vs. 8.5\%, adversarial vs. arbitrary), 2.2 times (15.0\% vs. 6.9\%), 4.4 times (71.1\% vs. 16.2\%), and 2.0 times (7.7\% vs. 3.8\%) the arbitrary model's decision for HELOC, COMPAS, Adult, and German respectively. The variation in Reconcile's success at returning decisions towards the reference model is likely due to differences in the dataset, such as prevalence of the minority class and dataset size. Though Reconcile may be successful in reversing disagreements towards the reference model, choosing an adversarial model reduces the chances of success in comparison to a random model.

\section{Current Laws and Regulations}
\label{sec:policy}

\begin{figure}[t]
\centering
\includegraphics[width=\textwidth]{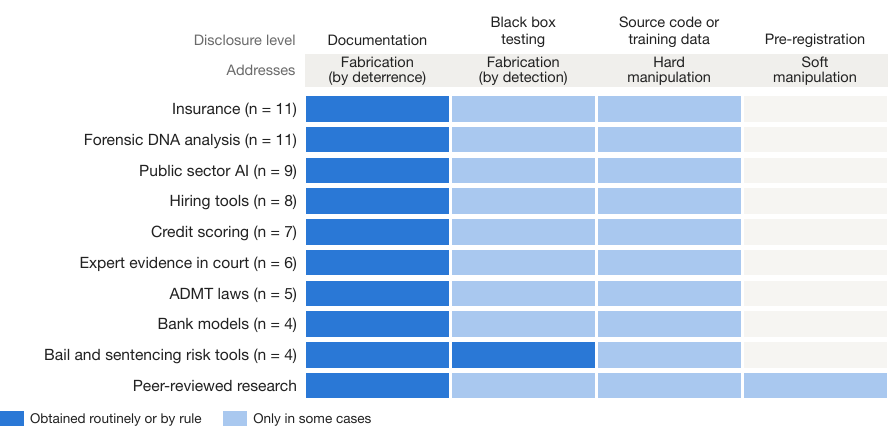}
\Description{Grid with ten sectors as rows (public sector AI, insurance, hiring tools, forensic DNA analysis, ADMT laws, expert evidence in court, credit scoring, bank models, bail and sentencing risk tools, peer-reviewed research) and four disclosure levels as columns (documentation, black box testing, source code or training data, pre-registration), each labelled with the manipulation it addresses.}
\caption{\textbf{Current regulation of machine learning and statistical models across sectors is not robust to adversarial multiplicity.} Indicators for what types of disclosure are routinely acquired or acquired only in some cases, across high-stakes settings. Disclosure level indicates the amount and type of information an auditor receives and what types of manipulation the information can identify.}
\label{fig:policy}
\end{figure}

We examined current regulation of machine learning and statistical models used for high-stakes decisions, and broadly find that adversarial multiplicity is not meaningfully addressed. 

Specifically, we conducted an LLM-based search starting from existing compilations of machine learning and regulations (including the 25 public-sector audit policies of \citet{escher2026algorithm}) to identify laws, regulations, guidance documents, and standards, followed by manual review, close reading, deductive coding for regulation categories, and abductive coding for thematic analysis of norms surrounding manipulation and bypassing regulatory oversight.

Figure~\ref{fig:policy} maps 54 enacted laws and regulations, guidance documents, and standards (plus 11 court rulings), split into the following categories. 
\textbf{Insurance}: insurers' pricing and claims models;
\textbf{Forensic DNA analysis}: software that statistically analyzes DNA mixtures in criminal cases;
\textbf{Public sector AI}: AI tools used or contracted by government agencies;
\textbf{Hiring tools}: automated screening or evaluation of job applicants;
\textbf{Credit scoring}: models used to decide consumer loans;
\textbf{Expert evidence in court}: general rules of discovery and evidence for when model outputs are used in court;
\textbf{ADMT laws}: laws governing automated decision-making tools;
\textbf{Bank models}: internal models used by banks to rate loan risk and set capital requirements;
\textbf{Bail and sentencing risk tools}: scoring models for risk of recidivism or missing court;
\textbf{Peer-reviewed research}: norms and requirements across scientific journals and conferences.

We found all settings require some form of documentation for their models, but typically few disclosure requirements beyond that. Consequently, current norms deter outright fabrication of model outputs as it may incur penalties for providing false statements, but would not catch soft or hard manipulation if documented in sufficiently broad or vague terms. In most settings, regulators or courts may request black box testing (testing how models respond to chosen inputs), and most allow requests or court orders for training data and/or source code. No settings require pre-registration---committing to a specific procedure or analysis plan prior to deployment---except where a researcher voluntarily does so for a study. Table~\ref{tab:policyoverview} in the Appendix lists each policy and ruling.

\subsection{Recommendations}
\label{sec:recs}

For cases involving individual analyses, such as in forensic DNA evidence or, more broadly, expert evidence, a simple fix for preventing adversarial multiplicity (or accusations thereof) is requiring pre-registration. Since in these cases, only one analysis is performed, pre-registering the search strategy for it is cheap and straightforward. Precedent exists in the form of FDA drug trials, where statistical analyses are required to be pre-registered \citep{ich1998e9, nosek2018}.

Deployed models face different challenges since they tend to be used in real-time production environments where models are continuously updated and redesigned to align with product development needs and goals. Pre-registration does not necessarily apply as a fix in these environments, as even objectives and outcome variables can plausibly change over time, which may be difficult to anticipate. The constant updating of models may also drive concerns of multiplicity: if a user contests one model output, can the dispute be judged by an entirely new model? Instead, a solution would be for firms to keep dated model versions, so contested decisions can be judged by the model that was used at the time of deployment---this would also naturally prevent firms from exploiting adversarial multiplicity to create a new advantageous model post hoc.

Other options include providing proof of training (in cases where data cannot be shared) or rerunning the training pipeline in order to prevent hard manipulation \citep{choi2023potd,waiwitlikhit2024zkaudit,melissaris2026zkboost}. Soft manipulation can also be mitigated by committing to versioned, publicly available, general-use models (such as various tabular foundation models \citep{hollmann2025, hollmann2022tabpfn}), that have fewer configurations to tweak, and therefore less room to leverage multiplicity.

\section{Discussion}
\label{sec:discussion}

Our results show that predictive multiplicity can be plausibly exploited to engineer desired outcomes for motivated parties, at low-to-no cost in predictive performance. We further find that various fixes, such as multi-metric evaluations, the Reconcile algorithm, or various existing disclosure requirements in regulatory environments, are broadly insufficient: some may mitigate the impact of adversarial multiplicity, but none outright remove the possibility. Instead, we recommend that disclosure environments acknowledge the possibility of adversarial multiplicity and design audit environments accordingly: for one-off analyses where it's feasible to pre-register a search space, do so; for models that are continuously deployed, require versioning of the models so disputes can't be assessed based on different model versions.

Our work is limited to the four public benchmark datasets commonly used in algorithmic fairness research, as well as the wildfire dataset---although we expect our theoretical results to generalize, the extent to which approaches like Reconcile successfully resolve disputes is context-dependent, based on information such as the prevalence of the minority class. Furthermore, our study primarily analyzes machine learning approaches on tabular datasets---more complex models like language or vision models may exhibit adversarial multiplicity differently. The modeling assumptions we use are generally conservative, such as holding all other predictions constant, or including manipulated observations in the scoring set---in practice, motivated actors would not need to do either of these and could simply let other observations change or exclude a targeted observation from the scoring set.

Is it likely that motivated actors would exploit adversarial multiplicity? Although it may be impossible to tell the extent to which it is or will be occurring, our work shows that the capacity exists to do so, and we argue that the incentive exists as well. In purely adversarial settings such as litigation or dispute resolution, parties are explicitly incentivized to select models that are advantageous for them. In other contexts, firms may choose models that are profit maximizing for them, such as banks selecting risk models that assign lower risk of default \citep{psi2013whale}. In such environments, oversight and clear procedural guidelines as to what constitutes proper model selection are necessary to prevent misuse. 

Hard manipulation may seem implausible given the effort required to do so, and the lack of plausible deniability that would otherwise be afforded by soft manipulation. However, looking forward, adversarial multiplicity may also end up inadvertently being misused by AI coding agents. As developers increasingly turn to agentic engineering (and therefore have less direct oversight of lines of code) \citep{chou2025vibe, becker2025}, it's plausible that coding agents gaming their objectives could simply induce hard or soft manipulation to obtain desired results. Therefore, malicious intent isn't even necessary for adversarial multiplicity to occur---in all of the examples and settings we analyze in this work, manipulation could occur either by a motivated party directly doing so, or by a developer inadvertently letting an agent do so without sufficient oversight. 

Solutions to detect and catch agent-driven adversarial multiplicity may involve more comprehensive code review, such as directing coding agents specifically to avoid adversarial multiplicity, or to explicitly consider how multiplicity may play a role in their model search. More speculatively, it's plausible that coding agents could even be used to conduct manipulations and then obfuscate their code afterwards to hide evidence of doing so---some works have found that auditors can only modestly catch sabotage of machine learning code \citep{gan2026, ward2025, libon2026}. As programming and modeling is increasingly delegated to agents, pre-registration remains a strong defense against adversarial multiplicity: a clear understanding beforehand of what the search space and objectives are, followed by a check if the final model actually came from the pre-specified search may serve as useful checks.

\bibliographystyle{plainnat}
\bibliography{refs}

\begin{thebibliography}{77}
\providecommand{\natexlab}[1]{#1}
\providecommand{\url}[1]{\texttt{#1}}
\expandafter\ifx\csname urlstyle\endcsname\relax
  \providecommand{\doi}[1]{doi: #1}\else
  \providecommand{\doi}{doi: \begingroup \urlstyle{rm}\Url}\fi

\bibitem[A{\"\i}vodji et~al.(2019)A{\"\i}vodji, Arai, Fortineau, Gambs, Hara,
  and Tapp]{aivodji2019}
Ulrich A{\"\i}vodji, Hiromi Arai, Olivier Fortineau, S{\'e}bastien Gambs,
  Satoshi Hara, and Alain Tapp.
\newblock Fairwashing: the risk of rationalization.
\newblock In \emph{International conference on machine learning}, pages
  161--170. PMLR, 2019.

\bibitem[Anders et~al.(2020)Anders, Pasliev, Dombrowski, M{\"u}ller, and
  Kessel]{anders2020fairwashing}
Christopher Anders, Plamen Pasliev, Ann-Kathrin Dombrowski, Klaus-Robert
  M{\"u}ller, and Pan Kessel.
\newblock Fairwashing explanations with off-manifold detergent.
\newblock In \emph{International Conference on Machine Learning}, pages
  314--323. PMLR, 2020.

\bibitem[Bai et~al.(2021)Bai, Wu, Zhang, Li, Li, and Xia]{bai2021}
Jiawang Bai, Baoyuan Wu, Yong Zhang, Yiming Li, Zhifeng Li, and Shu-Tao Xia.
\newblock Targeted attack against deep neural networks via flipping limited
  weight bits.
\newblock \emph{arXiv preprint arXiv:2102.10496}, 2021.

\bibitem[Becker et~al.(2025)Becker, Rush, Barnes, and Rein]{becker2025}
Joel Becker, Nate Rush, Elizabeth Barnes, and David Rein.
\newblock Measuring the impact of early-2025 ai on experienced open-source
  developer productivity.
\newblock \emph{arXiv preprint arXiv:2507.09089}, 2025.

\bibitem[Behzad et~al.(2025)Behzad, Casacuberta, Diana, and
  Tolbert]{behzad2025}
Tina Behzad, S{\'\i}lvia Casacuberta, Emily~Ruth Diana, and Alexander~Williams
  Tolbert.
\newblock Reconciling predictive multiplicity in practice.
\newblock In \emph{Proceedings of the 2025 ACM Conference on Fairness,
  Accountability, and Transparency}, pages 3350--3369, 2025.

\bibitem[Black et~al.(2022)Black, Raghavan, and Barocas]{black2022}
Emily Black, Manish Raghavan, and Solon Barocas.
\newblock Model multiplicity: Opportunities, concerns, and solutions.
\newblock In \emph{Proceedings of the 2022 ACM conference on fairness,
  accountability, and transparency}, pages 850--863, 2022.

\bibitem[Black et~al.(2024{\natexlab{a}})Black, Gillis, and Hall]{gillis2024}
Emily Black, Talia Gillis, and Zara~Yasmine Hall.
\newblock D-hacking.
\newblock In \emph{Proceedings of the 2024 ACM Conference on Fairness,
  Accountability, and Transparency}, pages 602--615, 2024{\natexlab{a}}.

\bibitem[Black et~al.(2024{\natexlab{b}})Black, Koepke, Kim, Barocas, and
  Hsu]{blackldalaw2024}
Emily Black, John~Logan Koepke, Pauline~T Kim, Solon Barocas, and Mingwei Hsu.
\newblock Less discriminatory algorithms.
\newblock \emph{Geo. LJ}, 113:\penalty0 53, 2024{\natexlab{b}}.

\bibitem[Bourr{\'e}e et~al.(2025)Bourr{\'e}e, Godinot, De~Vos, Vujasinovic,
  Biswas, Tredan, Merrer, and Kermarrec]{garciabourree2025robust}
Jade~Garcia Bourr{\'e}e, Augustin Godinot, Martijn De~Vos, Milos Vujasinovic,
  Sayan Biswas, Gilles Tredan, Erwan~Le Merrer, and Anne-Marie Kermarrec.
\newblock Robust ml auditing using prior knowledge.
\newblock \emph{arXiv preprint arXiv:2505.04796}, 2025.

\bibitem[Breiman(2001{\natexlab{a}})]{breiman2001}
Leo Breiman.
\newblock Statistical modeling: The two cultures (with comments and a rejoinder
  by the author).
\newblock \emph{Statistical science}, 16\penalty0 (3):\penalty0 199--231,
  2001{\natexlab{a}}.

\bibitem[Breiman(2001{\natexlab{b}})]{breiman2001random}
Leo Breiman.
\newblock Random forests.
\newblock \emph{Machine learning}, 45\penalty0 (1):\penalty0 5--32,
  2001{\natexlab{b}}.

\bibitem[Brown et~al.(2023)Brown, Hanley, Mahesh, Reed, Strenfel, Davis,
  Kochanski, and Clements]{brown2023}
Patrick~T Brown, Holt Hanley, Ankur Mahesh, Colorado Reed, Scott~J Strenfel,
  Steven~J Davis, Adam~K Kochanski, and Craig~B Clements.
\newblock Climate warming increases extreme daily wildfire growth risk in
  california.
\newblock \emph{Nature}, 621\penalty0 (7980):\penalty0 760--766, 2023.

\bibitem[Burger et~al.(2019)Burger, Horton, and Wentz]{burger2019law}
Michael Burger, Radley Horton, and Jessica Wentz.
\newblock The law and science of climate change attribution.
\newblock \emph{Forthcoming, Columbia Journal of Environmental Law (January
  2020)}, 2019.

\bibitem[Casper et~al.(2024)Casper, Ezell, Siegmann, Kolt, Curtis, Bucknall,
  Haupt, Wei, Scheurer, Hobbhahn, et~al.]{casper2024blackbox}
Stephen Casper, Carson Ezell, Charlotte Siegmann, Noam Kolt, Taylor~Lynn
  Curtis, Benjamin Bucknall, Andreas Haupt, Kevin Wei, J{\'e}r{\'e}my Scheurer,
  Marius Hobbhahn, et~al.
\newblock Black-box access is insufficient for rigorous ai audits.
\newblock In \emph{Proceedings of the 2024 ACM Conference on Fairness,
  Accountability, and Transparency}, pages 2254--2272, 2024.

\bibitem[Chen and Guestrin(2016)]{chen2016xgboost}
Tianqi Chen and Carlos Guestrin.
\newblock Xgboost: A scalable tree boosting system.
\newblock In \emph{Proceedings of the 22nd acm sigkdd international conference
  on knowledge discovery and data mining}, pages 785--794, 2016.

\bibitem[Choi et~al.(2023)Choi, Shavit, and Duvenaud]{choi2023potd}
Dami Choi, Yonadav Shavit, and David~K Duvenaud.
\newblock Tools for verifying neural models' training data.
\newblock \emph{Advances in Neural Information Processing Systems},
  36:\penalty0 1154--1188, 2023.

\bibitem[Chou et~al.(2025)Chou, Zeger, and Huynh]{chou2026}
Cassandra~C Chou, Scott~L Zeger, and Benjamin~Q Huynh.
\newblock Validity in machine learning for extreme event attribution.
\newblock \emph{arXiv preprint arXiv:2511.19039}, 2025.

\bibitem[Chou et~al.(2026)Chou, Jiang, Chen, Weng, Jackson, Zimmermann, and
  Jones]{chou2025vibe}
Yi-Hung Chou, Boyuan Jiang, Yi~Wen Chen, Mingyue Weng, Victoria Jackson, Thomas
  Zimmermann, and James~A Jones.
\newblock Building software by rolling the dice: A qualitative study of vibe
  coding.
\newblock \emph{Proceedings of the ACM on Software Engineering}, 3\penalty0
  (FSE):\penalty0 1712--1734, 2026.

\bibitem[Cooper et~al.(2021)Cooper, Lu, Forde, and De~Sa]{cooper2021hpo}
A~Feder Cooper, Yucheng Lu, Jessica Forde, and Christopher~M De~Sa.
\newblock Hyperparameter optimization is deceiving us, and how to stop it.
\newblock \emph{Advances in Neural Information Processing Systems},
  34:\penalty0 3081--3095, 2021.

\bibitem[Coston et~al.(2021)Coston, Rambachan, and
  Chouldechova]{coston2021fairs}
Amanda Coston, Ashesh Rambachan, and Alexandra Chouldechova.
\newblock Characterizing fairness over the set of good models under selective
  labels.
\newblock In \emph{International Conference on Machine Learning}, pages
  2144--2155. PMLR, 2021.

\bibitem[Dai et~al.(2025)Dai, Ravishankar, Yuan, Black, and Neill]{dai2025}
Gordon Dai, Pavan Ravishankar, Rachel Yuan, Emily Black, and Daniel~B Neill.
\newblock Be intentional about fairness!: Fairness, size, and multiplicity in
  the rashomon set.
\newblock In \emph{Proceedings of the 5th ACM Conference on Equity and Access
  in Algorithms, Mechanisms, and Optimization}, pages 42--73, 2025.

\bibitem[D'Amour et~al.(2022)D'Amour, Heller, Moldovan, Adlam, Alipanahi,
  Beutel, Chen, Deaton, Eisenstein, Hoffman, et~al.]{damour2022}
Alexander D'Amour, Katherine Heller, Dan Moldovan, Ben Adlam, Babak Alipanahi,
  Alex Beutel, Christina Chen, Jonathan Deaton, Jacob Eisenstein, Matthew~D
  Hoffman, et~al.
\newblock Underspecification presents challenges for credibility in modern
  machine learning.
\newblock \emph{Journal of Machine Learning Research}, 23\penalty0
  (226):\penalty0 1--61, 2022.

\bibitem[Davis et~al.(2022)Davis, Lo, Mishra, Nourian, Singh, Wu, and
  Zhang]{davis2022explainable}
Randall Davis, Andrew~W Lo, Sudhanshu Mishra, Arash Nourian, Manish Singh,
  Nicholas Wu, and Ruixun Zhang.
\newblock Explainable machine learning models of consumer credit risk.
\newblock \emph{Available at SSRN}, 4006840, 2022.

\bibitem[Du et~al.(2025)Du, Ngo, and Wu]{du2024}
Ally Du, Dung~Daniel Ngo, and Steven Wu.
\newblock Reconciling model multiplicity for downstream decision making.
\newblock In \emph{International Conference on Learning Representations},
  volume 2025, pages 59133--59164, 2025.

\bibitem[Dumford and Scheirer(2020)]{dumford2020}
Jacob Dumford and Walter Scheirer.
\newblock Backdooring convolutional neural networks via targeted weight
  perturbations.
\newblock In \emph{2020 IEEE International Joint Conference on Biometrics
  (IJCB)}, pages 1--9. IEEE, 2020.

\bibitem[{Equal Employment Opportunity Commission and Civil Service Commission
  and Department of Labor and Department of
  Justice}(1978)]{uniformguidelines1978}
{Equal Employment Opportunity Commission and Civil Service Commission and
  Department of Labor and Department of Justice}.
\newblock Uniform guidelines on employee selection procedures.
\newblock 43 Fed. Reg. 38295 (Aug. 25, 1978), codified at 29 C.F.R. Part 1607,
  1978.
\newblock Four-fifths rule at \S 1607.4(D).

\bibitem[Escher et~al.(2026)Escher, Banovic, and Green]{escher2026algorithm}
Nel Escher, Nikola Banovic, and Ben Green.
\newblock Algorithm auditing policies rest on flawed assumptions about public
  sector systems.
\newblock In \emph{The 2026 ACM Conference on Fairness, Accountability, and
  Transparency}, pages 2162--2185, 2026.

\bibitem[Fisher et~al.(2019)Fisher, Rudin, and Dominici]{fisher2019all}
Aaron Fisher, Cynthia Rudin, and Francesca Dominici.
\newblock All models are wrong, but many are useful: Learning a variable's
  importance by studying an entire class of prediction models simultaneously.
\newblock \emph{Journal of machine learning research}, 20\penalty0
  (177):\penalty0 1--81, 2019.

\bibitem[Gan et~al.(2026)Gan, Bhatt, Shlegeris, Stastny, and Hebbar]{gan2026}
Eric Gan, Aryan Bhatt, Buck Shlegeris, Julian Stastny, and Vivek Hebbar.
\newblock Auditing sabotage bench: A benchmark for detecting and fixing
  research sabotage in ml codebases.
\newblock \emph{arXiv preprint arXiv:2604.16286}, 2026.

\bibitem[Ganesh et~al.(2023)Ganesh, Chang, Strobel, and
  Shokri]{ganesh2023randomness}
Prakhar Ganesh, Hongyan Chang, Martin Strobel, and Reza Shokri.
\newblock On the impact of machine learning randomness on group fairness.
\newblock In \emph{Proceedings of the 2023 ACM Conference on Fairness,
  Accountability, and Transparency}, pages 1789--1800, 2023.

\bibitem[Ganesh et~al.(2025)Ganesh, Taik, and Farnadi]{ganesh2025}
Prakhar Ganesh, Afaf Taik, and Golnoosh Farnadi.
\newblock Systemizing multiplicity: The curious case of arbitrariness in
  machine learning.
\newblock In \emph{Proceedings of the AAAI/ACM Conference on AI, Ethics, and
  Society}, volume~8, pages 1032--1048, 2025.

\bibitem[Godinot et~al.(2024)Godinot, Le~Merrer, Tr{\'e}dan, Penzo, and
  Ta{\"\i}ani]{godinot2024harder}
Augustin Godinot, Erwan Le~Merrer, Gilles Tr{\'e}dan, Camilla Penzo, and
  Franois Ta{\"\i}ani.
\newblock Under manipulations, are some ai models harder to audit?
\newblock In \emph{2024 IEEE Conference on Secure and Trustworthy Machine
  Learning (SaTML)}, pages 644--664. IEEE, 2024.

\bibitem[Guideline(1999)]{ich1998e9}
ICH Harmonised~Tripartite Guideline.
\newblock Statistical principles for clinical trials. international conference
  on harmonisation e9 expert working group.
\newblock \emph{Stat. Med.}, 18:\penalty0 1905--1942, 1999.

\bibitem[Hofmann(1994)]{statlog_german_credit_data_144}
Hans Hofmann.
\newblock {Statlog (German Credit Data)}.
\newblock UCI Machine Learning Repository, 1994.
\newblock {DOI}: https://doi.org/10.24432/C5NC77.

\bibitem[Hollmann et~al.(2022)Hollmann, M{\"u}ller, Eggensperger, and
  Hutter]{hollmann2022tabpfn}
Noah Hollmann, Samuel M{\"u}ller, Katharina Eggensperger, and Frank Hutter.
\newblock Tabpfn: A transformer that solves small tabular classification
  problems in a second.
\newblock \emph{arXiv preprint arXiv:2207.01848}, 2022.

\bibitem[Hollmann et~al.(2025)Hollmann, M{\"u}ller, Purucker, Krishnakumar,
  K{\"o}rfer, Hoo, Schirrmeister, and Hutter]{hollmann2025}
Noah Hollmann, Samuel M{\"u}ller, Lennart Purucker, Arjun Krishnakumar, Max
  K{\"o}rfer, Shi~Bin Hoo, Robin~Tibor Schirrmeister, and Frank Hutter.
\newblock Accurate predictions on small data with a tabular foundation model.
\newblock \emph{Nature}, 637\penalty0 (8045):\penalty0 319--326, 2025.

\bibitem[Hong et~al.(2022)Hong, Carlini, and Kurakin]{hong2022}
Sanghyun Hong, Nicholas Carlini, and Alexey Kurakin.
\newblock Handcrafted backdoors in deep neural networks.
\newblock \emph{Advances in Neural Information Processing Systems},
  35:\penalty0 8068--8080, 2022.

\bibitem[Hsu and Calmon(2022)]{hsu2022}
Hsiang Hsu and Flavio Calmon.
\newblock Rashomon capacity: A metric for predictive multiplicity in
  classification.
\newblock \emph{Advances in Neural Information Processing Systems},
  35:\penalty0 28988--29000, 2022.

\bibitem[Hsu et~al.(2024)Hsu, Li, Hu, and Chen]{hsu2024}
Hsiang Hsu, Guihong Li, Shaohan Hu, and Chun-Fu Chen.
\newblock Dropout-based rashomon set exploration for efficient predictive
  multiplicity estimation.
\newblock In \emph{International Conference on Learning Representations},
  volume 2024, pages 33104--33144, 2024.

\bibitem[Jia et~al.(2021)Jia, Yaghini, Choquette-Choo, Dullerud, Thudi,
  Chandrasekaran, and Papernot]{jia2021pol}
Hengrui Jia, Mohammad Yaghini, Christopher~A Choquette-Choo, Natalie Dullerud,
  Anvith Thudi, Varun Chandrasekaran, and Nicolas Papernot.
\newblock Proof-of-learning: Definitions and practice.
\newblock In \emph{2021 IEEE Symposium on Security and Privacy (SP)}, pages
  1039--1056. IEEE, 2021.

\bibitem[Ke et~al.(2017)Ke, Meng, Finley, Wang, Chen, Ma, Ye, and
  Liu]{ke2017lightgbm}
Guolin Ke, Qi~Meng, Thomas Finley, Taifeng Wang, Wei Chen, Weidong Ma, Qiwei
  Ye, and Tie-Yan Liu.
\newblock Lightgbm: A highly efficient gradient boosting decision tree.
\newblock \emph{Advances in neural information processing systems}, 30, 2017.

\bibitem[Kroll(2015)]{kroll2017}
Joshua~Alexander Kroll.
\newblock \emph{Accountable algorithms}.
\newblock Princeton University, 2015.

\bibitem[Larson et~al.(2016)Larson, Mattu, Kirchner, and Angwin]{larson2016we}
Jeff Larson, Surya Mattu, Lauren Kirchner, and Julia Angwin.
\newblock How we analyzed the compas recidivism algorithm.
\newblock \emph{ProPublica (5 2016)}, 9\penalty0 (1):\penalty0 3--3, 2016.

\bibitem[Le~Merrer and Tr{\'e}dan(2020)]{lemerrer2020bouncer}
Erwan Le~Merrer and Gilles Tr{\'e}dan.
\newblock Remote explainability faces the bouncer problem.
\newblock \emph{Nature machine intelligence}, 2\penalty0 (9):\penalty0
  529--539, 2020.

\bibitem[Libon et~al.(2026)Libon, Rank, Yeon, Schmotz, Qin, Donnelly,
  Prinzhorn, and Andriushchenko]{libon2026}
Lena Libon, Ben Rank, Jehyeok Yeon, David Schmotz, Jeremy Qin, Daniel Donnelly,
  Derck Prinzhorn, and Maksym Andriushchenko.
\newblock Researcharena: Evaluating sabotage and monitoring in automated ai
  r\&d.
\newblock \emph{arXiv preprint arXiv:2607.19321}, 2026.

\bibitem[Lloyd and Shepherd(2021)]{lloyd2021climate}
Elisabeth~A Lloyd and Theodore~G Shepherd.
\newblock Climate change attribution and legal contexts: evidence and the role
  of storylines.
\newblock \emph{Climatic Change}, 167\penalty0 (3):\penalty0 28, 2021.

\bibitem[Lloyd et~al.(2021)Lloyd, Oreskes, Seneviratne, and
  Larson]{lloyd2021proof}
Elisabeth~A Lloyd, Naomi Oreskes, Sonia~I Seneviratne, and Edward~J Larson.
\newblock Climate scientists set the bar of proof too high.
\newblock \emph{Climatic Change}, 165\penalty0 (3):\penalty0 55, 2021.

\bibitem[Lucena(2018)]{lucena2018spline}
Brian Lucena.
\newblock Spline-based probability calibration.
\newblock \emph{arXiv preprint arXiv:1809.07751}, 2018.

\bibitem[Luck et~al.(2026)Luck, Franzese-McLaughlin, Masserova, Takahashi,
  Polychroniadou, and Papernot]{luck2026cmc}
Carter Luck, Olive Franzese-McLaughlin, Elisaweta Masserova, Akira Takahashi,
  Antigoni Polychroniadou, and Nicolas Papernot.
\newblock Certified in theory, broken in practice: Assumption gaps in
  cryptographic model certification.
\newblock In \emph{35th USENIX Security Symposium (USENIX Security 26)}, pages
  6007--6026, 2026.

\bibitem[Marx et~al.(2020)Marx, Calmon, and Ustun]{marx2020}
Charles Marx, Flavio Calmon, and Berk Ustun.
\newblock Predictive multiplicity in classification.
\newblock In \emph{International conference on machine learning}, pages
  6765--6774. PMLR, 2020.

\bibitem[Meding and Hagendorff(2024)]{meding2024fairnesshacking}
Kristof Meding and Thilo Hagendorff.
\newblock Fairness hacking: The malicious practice of shrouding unfairness in
  algorithms: K. meding et al.
\newblock \emph{Philosophy \& Technology}, 37\penalty0 (1):\penalty0 4, 2024.

\bibitem[Melissaris et~al.(2026)Melissaris, Polychroniadou, Takahashi, Weng,
  and Xu]{melissaris2026zkboost}
Nikolas Melissaris, Antigoni Polychroniadou, Akira Takahashi, Chenkai Weng, and
  Jiayi Xu.
\newblock Zkboost: Zero-knowledge verifiable training for xgboost.
\newblock \emph{arXiv preprint arXiv:2602.04113}, 2026.

\bibitem[{New York City Department of Consumer and Worker
  Protection}(2023)]{nycdcwp2023aedt}
{New York City Department of Consumer and Worker Protection}.
\newblock Notice of adoption of final rule: Automated employment decision
  tools.
\newblock 6 RCNY \S\S 5-300--5-304, adopted April 6, 2023, 2023.
\newblock Implementing Local Law 144 of 2021.

\bibitem[Niculescu-Mizil and Caruana(2005)]{niculescu2005obtaining}
Alexandru Niculescu-Mizil and Rich Caruana.
\newblock Obtaining calibrated probabilities from boosting.
\newblock In \emph{UAI}, volume~5, pages 413--20, 2005.

\bibitem[Nosek et~al.(2018)Nosek, Ebersole, DeHaven, and Mellor]{nosek2018}
Brian~A Nosek, Charles~R Ebersole, Alexander~C DeHaven, and David~T Mellor.
\newblock The preregistration revolution.
\newblock \emph{Proceedings of the National Academy of Sciences}, 115\penalty0
  (11):\penalty0 2600--2606, 2018.

\bibitem[Pedregosa et~al.(2011)Pedregosa, Varoquaux, Gramfort, Michel, Thirion,
  Grisel, Blondel, Prettenhofer, Weiss, Dubourg, et~al.]{pedregosa2011scikit}
Fabian Pedregosa, Ga{\"e}l Varoquaux, Alexandre Gramfort, Vincent Michel,
  Bertrand Thirion, Olivier Grisel, Mathieu Blondel, Peter Prettenhofer, Ron
  Weiss, Vincent Dubourg, et~al.
\newblock Scikit-learn: Machine learning in python.
\newblock \emph{the Journal of machine Learning research}, 12:\penalty0
  2825--2830, 2011.

\bibitem[Philip et~al.(2020)Philip, Kew, van Oldenborgh, Otto, Vautard, van
  Der~Wiel, King, Lott, Arrighi, Singh, et~al.]{philip2020}
Sjoukje Philip, Sarah Kew, Geert~Jan van Oldenborgh, Friederike Otto, Robert
  Vautard, Karin van Der~Wiel, Andrew King, Fraser Lott, Julie Arrighi, Roop
  Singh, et~al.
\newblock A protocol for probabilistic extreme event attribution analyses.
\newblock \emph{Advances in Statistical Climatology, Meteorology and
  Oceanography}, 6\penalty0 (2):\penalty0 177--203, 2020.

\bibitem[Platt et~al.(1999)]{platt1999probabilistic}
John Platt et~al.
\newblock Probabilistic outputs for support vector machines and comparisons to
  regularized likelihood methods.
\newblock \emph{Advances in large margin classifiers}, 10\penalty0
  (3):\penalty0 61--74, 1999.

\bibitem[Prokhorenkova et~al.(2018)Prokhorenkova, Gusev, Vorobev, Dorogush, and
  Gulin]{prokhorenkova2018catboost}
Liudmila Prokhorenkova, Gleb Gusev, Aleksandr Vorobev, Anna~Veronika Dorogush,
  and Andrey Gulin.
\newblock Catboost: unbiased boosting with categorical features.
\newblock \emph{Advances in neural information processing systems}, 31, 2018.

\bibitem[Ronny and Barry(1996)]{ronny1996uci}
Kohavi Ronny and Becker Barry.
\newblock Uci machine learning repository: Adult data set.
\newblock \emph{Accessed: Apr}, 30, 1996.

\bibitem[Roth et~al.(2023)Roth, Tolbert, and Weinstein]{roth2023}
Aaron Roth, Alexander Tolbert, and Scott Weinstein.
\newblock Reconciling individual probability forecasts.
\newblock In \emph{Proceedings of the 2023 ACM Conference on Fairness,
  Accountability, and Transparency}, pages 101--110, 2023.

\bibitem[Rumelhart et~al.(1986)Rumelhart, Hinton, and
  Williams]{rumelhart1986learning}
David~E Rumelhart, Geoffrey~E Hinton, and Ronald~J Williams.
\newblock Learning representations by back-propagating errors.
\newblock \emph{nature}, 323\penalty0 (6088):\penalty0 533--536, 1986.

\bibitem[Semenova et~al.(2022)Semenova, Rudin, and Parr]{semenova2022}
Lesia Semenova, Cynthia Rudin, and Ronald Parr.
\newblock On the existence of simpler machine learning models.
\newblock In \emph{Proceedings of the 2022 ACM Conference on Fairness,
  Accountability, and Transparency}, FAccT '22, page 1827–1858, New York, NY,
  USA, 2022. Association for Computing Machinery.
\newblock ISBN 9781450393522.
\newblock \doi{10.1145/3531146.3533232}.
\newblock URL \url{https://doi.org/10.1145/3531146.3533232}.

\bibitem[Shahin~Shamsabadi et~al.(2022)Shahin~Shamsabadi, Yaghini, Dullerud,
  Wyllie, A{\"\i}vodji, Alaagib, Gambs, and Papernot]{shamsabadi2022washing}
Ali Shahin~Shamsabadi, Mohammad Yaghini, Natalie Dullerud, Sierra Wyllie,
  Ulrich A{\"\i}vodji, Aisha Alaagib, S{\'e}bastien Gambs, and Nicolas
  Papernot.
\newblock Washing the unwashable: On the (im) possibility of fairwashing
  detection.
\newblock \emph{Advances in Neural Information Processing Systems},
  35:\penalty0 14170--14182, 2022.

\bibitem[Sharma et~al.(2024)Sharma, Redyuk, Mukherjee, {\v{S}}ipka,
  H{\"u}llermeier, Vollmer, and Selby]{sharma2025}
Rahul Sharma, Sergey Redyuk, Sumantrak Mukherjee, Andrea {\v{S}}ipka, Eyke
  H{\"u}llermeier, Sebastian Vollmer, and David Selby.
\newblock X hacking: The threat of misguided automl.
\newblock \emph{arXiv preprint arXiv:2401.08513}, 2024.

\bibitem[Silberzahn et~al.(2018)Silberzahn, Uhlmann, Martin, Anselmi, Aust,
  Awtrey, Bahn{\'\i}k, Bai, Bannard, Bonnier, et~al.]{silberzahn2018}
Raphael Silberzahn, Eric~L Uhlmann, Daniel~P Martin, Pasquale Anselmi, Frederik
  Aust, Eli Awtrey, {\v{S}}t{\v{e}}p{\'a}n Bahn{\'\i}k, Feng Bai, Colin
  Bannard, Evelina Bonnier, et~al.
\newblock Many analysts, one data set: Making transparent how variations in
  analytic choices affect results.
\newblock \emph{Advances in methods and practices in psychological science},
  1\penalty0 (3):\penalty0 337--356, 2018.

\bibitem[Simonsohn et~al.(2020)Simonsohn, Simmons, and Nelson]{simonsohn2020}
Uri Simonsohn, Joseph~P Simmons, and Leif~D Nelson.
\newblock Specification curve analysis.
\newblock \emph{Nature human behaviour}, 4\penalty0 (11):\penalty0 1208--1214,
  2020.

\bibitem[Simson et~al.(2024)Simson, Pfisterer, and Kern]{simson2024onemodel}
Jan Simson, Florian Pfisterer, and Christoph Kern.
\newblock One model many scores: Using multiverse analysis to prevent fairness
  hacking and evaluate the influence of model design decisions.
\newblock In \emph{Proceedings of the 2024 ACM Conference on Fairness,
  Accountability, and Transparency}, pages 1305--1320, 2024.

\bibitem[Steegen et~al.(2016)Steegen, Tuerlinckx, Gelman, and
  Vanpaemel]{steegen2016}
Sara Steegen, Francis Tuerlinckx, Andrew Gelman, and Wolf Vanpaemel.
\newblock Increasing transparency through a multiverse analysis.
\newblock \emph{Perspectives on Psychological Science}, 11\penalty0
  (5):\penalty0 702--712, 2016.

\bibitem[Stuart-Smith et~al.(2021)Stuart-Smith, Otto, Saad, Lisi, Minnerop,
  Lauta, Van~Zwieten, and Wetzer]{stuartsmith2021}
Rupert~F Stuart-Smith, Friederike~EL Otto, Aisha~I Saad, Gaia Lisi, Petra
  Minnerop, Kristian~Cedervall Lauta, Kristin Van~Zwieten, and Thom Wetzer.
\newblock Filling the evidentiary gap in climate litigation.
\newblock \emph{Nature Climate Change}, 11\penalty0 (8):\penalty0 651--655,
  2021.

\bibitem[Trok et~al.(2024)Trok, Barnes, Davenport, and Diffenbaugh]{trok2024}
Jared~T Trok, Elizabeth~A Barnes, Frances~V Davenport, and Noah~S Diffenbaugh.
\newblock Machine learning--based extreme event attribution.
\newblock \emph{Science Advances}, 10\penalty0 (34):\penalty0 eadl3242, 2024.

\bibitem[{U.S. Senate Permanent Subcommittee on
  Investigations}(2013)]{psi2013whale}
{U.S. Senate Permanent Subcommittee on Investigations}.
\newblock {JPMorgan Chase} whale trades: A case history of derivatives risks
  and abuses.
\newblock Majority and minority staff report, Committee on Homeland Security
  and Governmental Affairs, United States Senate, March 2013.
\newblock URL
  \url{https://www.hsgac.senate.gov/wp-content/uploads/imo/media/doc/REPORT%20-%20JPMorgan%20Chase%20Whale%20Trades%20(3-15-13).pdf}.
\newblock Released in conjunction with the March 15, 2013 hearing.

\bibitem[Waiwitlikhit et~al.(2024)Waiwitlikhit, Stoica, Sun, Hashimoto, and
  Kang]{waiwitlikhit2024zkaudit}
Suppakit Waiwitlikhit, Ion Stoica, Yi~Sun, Tatsunori Hashimoto, and Daniel
  Kang.
\newblock Trustless audits without revealing data or models.
\newblock \emph{arXiv preprint arXiv:2404.04500}, 2024.

\bibitem[Ward et~al.(2026)Ward, van~der Weij, G{\'a}bor, Martin, Moreno, Lidar,
  Makower, Jodrell, and Robson]{ward2025}
Francis Ward, Teun van~der Weij, Hanna G{\'a}bor, Sam Martin, Raja Moreno,
  Harel Lidar, Louis Makower, Thomas Jodrell, and Lauren Robson.
\newblock Ctrl-alt-deceit sabotage evaluations for automated ai r\&d.
\newblock \emph{Advances in Neural Information Processing Systems},
  38:\penalty0 173596--173648, 2026.

\bibitem[Watson-Daniels et~al.(2023)Watson-Daniels, Parkes, and
  Ustun]{watsondaniels2023}
Jamelle Watson-Daniels, David~C Parkes, and Berk Ustun.
\newblock Predictive multiplicity in probabilistic classification.
\newblock In \emph{Proceedings of the AAAI Conference on Artificial
  Intelligence}, volume~37, pages 10306--10314, 2023.

\bibitem[Yan and Zhang(2022)]{yan2022active}
Tom Yan and Chicheng Zhang.
\newblock Active fairness auditing.
\newblock In \emph{International Conference on Machine Learning}, pages
  24929--24962. PMLR, 2022.

\bibitem[Zhao et~al.(2019)Zhao, Wang, Gongye, Wang, Fei, and Lin]{zhao2019}
Pu~Zhao, Siyue Wang, Cheng Gongye, Yanzhi Wang, Yunsi Fei, and Xue Lin.
\newblock Fault sneaking attack: A stealthy framework for misleading deep
  neural networks.
\newblock In \emph{Proceedings of the 56th Annual Design Automation Conference
  2019}, pages 1--6, 2019.

\end{thebibliography}

\appendix

\section{Proof of Theorem~\ref{thm:hard}}
\label{app:hard}

\begin{proof}
Let $x$ be the scored data point intended to be manipulated, and let $x'$ be any other scored data point not functionally equivalent to $x$, such that $\phi(x) \neq \phi(x')$. Consider that any changing of parameters $\theta$ by $d$ changes the raw outputs $\theta^\top\phi(x')$ by $d^\top\phi(x')$. Consider also that the function $\psi$ converts the model's raw output $\theta^\top\phi(x')$ into its prediction (e.g., a sigmoid function converting log-odds into a probability). Because $\psi$ is strictly increasing, fixing the predictions will also fix the raw outputs. Therefore, we need to find some $d$ that changes $f(x)$ while holding $f(x')$ constant. Formally, for any shift $\Delta$ in the raw output of $x$, we search for $d$ with $d^\top\phi(x)=\Delta$ and $d^\top\phi(x')=0$ for every $x'$.

\emph{If.} Let $V$ be the set of distinct vectors $\phi(\cdot)$ of the scored data points. When the vectors in $V$ are linearly independent, $\phi(x)$ is not in the span of the $\phi(x')$, so its component $w$ orthogonal to that span is nonzero, with $w^\top\phi(x')=0$ and $w^\top\phi(x)=\|w\|^2$. It follows that $d=\Delta\,w/\|w\|^2$ applies for every $\Delta$.

\emph{Only if.} When the vectors in $V$ are linearly dependent, for some scored data point $x$ we have $\phi(x)=\sum_{x'}b_{x'}\phi(x')$. Any $d$ that keeps every $f(x')$ fixed has $d^\top\phi(x')=0$ for every $x'$, so $d^\top\phi(x)=\sum_{x'}b_{x'}\,d^\top\phi(x')=0$, and $f(x)$ cannot change.
\end{proof}

\section{Methods}
\label{app:methods}
\subsection{Data}
\label{supp:data}
We use four datasets that are frequently used in fair machine learning: HELOC, COMPAS, Adult, and German. We additionally use California wildfire data from 2003--2020 from \citet{brown2023}.

HELOC has 10,000 applicants and predicts whether credit applicants to FICO's home-equity line will be 90 or more days late on a payment within 24 months. We use 19 additional predictors, which cover the following areas: risk score from an outside model, credit history (months since the oldest and newest account was opened, average months the accounts have been on file, number of accounts opened in the last 12 months), credit payment history (number of accounts in good standing, number of accounts with 2+ payments at least 60 days late, number of accounts that never had late payments, months since the most recent late payment, worst credit status in the last 12 months by FICO coding), mix of credit (share of accounts that are installment loans), recent credit inquiries (months since last credit inquiry, number of inquiries in the last 6 months), and debt load (revolving balance as a percentage of credit limit, installment balance as a percentage of loan amount, number of revolving accounts carrying a balance, number of installment accounts carrying a balance, number of bank or national-card accounts that are near their limits, percentage of accounts carrying a balance).

COMPAS has 5,278 defendants, and predicts whether defendants will be re-arrested within two years. We use the following predictors: sex, age, age category, race, whether the individual has at least three juvenile offense counts, count of prior offenses, and charge degree.

Adult has 48,842 individuals, and the outcome of interest is whether an individual has income above \$50,000. We use the following predictors, where we use one-hot encoding for categorical variables: age in years, census sampling weight, education level, investment gains and losses, average number of hours worked per week, country of birth, work class (private, government, self-employed, without pay, never worked), education category, marital status, occupation category (e.g., executive, professional specialty, sales, etc.), role in the household, race, and sex.

German credit has 1,000 applicants, and the binary outcome of interest is whether an individual is a good or bad credit risk. We use 20 predictors, and categorical values are one-hot encoded: loan length in months, loan amount, installment as a share of disposable income, years of residence at the current address, age in years, number of existing loans at this bank, number of applicant's dependents, checking-account balance (below 0, 0–200 DM, 200 DM or more, no account), credit history category (all credits paid back duly, existing credits paid duly, delays in the past, critical account or credits elsewhere), purpose of loan, savings status (below 100, 100–500, 500–1,000, 1,000 DM or more, no known savings), years in the current job (unemployed, under 1, 1–4, 4–7, 7 or more), sex and marital status, other parties on the loan, most valuable property, other installment plans, housing category (own, rent, live for free), job category (unemployed or unskilled non-resident, unskilled resident, skilled, highly qualified/self-employed/management), binary indicator of telephone ownership, and binary indicator of foreign worker status.

California wildfire data includes 17,910 fire days from 2003--2020, 380 of which were extreme fire growth days where the fire grew more than 10,000 acres in a single day, and the outcome of interest is extreme fire growth. The predictor variables include the following: temperature, vapor pressure deficit, 100 hour dead fuel moisture and 1000 hour dead fuel moisture, precipitation, wind speed, land use category (forest, shrub, or savanna/grassland), land aspect (cardinal direction), land slope, elevation above sea level, and vegetation fraction. We have observed data from the present-day, and simulated data from the pre-industrial time period. As described in \citet{brown2023}, temperature is changed using general circulation models and propagated into vapor pressure deficit and 100- and 1000-hour dead fuel moisture using physics-based equations to represent a pre-industrial time period. The rest of the predictor variables remain the same such that each day in the present-day dataset has a matching day in the pre-industrial time period.

\subsection{Hard manipulation: model training}
We use HELOC, COMPAS, Adult, and German with a 60/20/20 training/validation/test split (seed 0) such that HELOC, COMPAS, Adult, and German have 2,000, 1,055, 9,768, and 200 validation rows respectively. Models are fit on the training data and scored based on the validation data. The decision threshold is $t = 0.5$. We use one model family: LightGBM. We chose LightGBM because it is the state-of-the-art tree-ensemble model, which allows for exact manipulation of the leaf values.

To identify our reference model, we use 5-fold cross-validated log loss on the training data. We choose the best model from 72 configurations per dataset (learning rate 0.01, 0.03; 31, 63, 127 leaves; minimum leaf size 1, 5, 20, 100; L2 penalty 1, 10, 100). We define the Rashomon set as the set of models with a validation Brier score within $\varepsilon = 1\%$ of the validation Brier of the reference model, which results in 65 models in the Rashomon set for HELOC, 54 models for COMPAS, 62 models for German, and 39 models for Adult. Brier score is the mean squared difference between predicted probabilities and true outcomes, and can be represented as: $\text{Brier Score} = \frac{1}{n} \sum_{i=1}^{n} (\hat{p}(X_i^{(0)})-Y_i^{(0)})^2$. To identify the starting model (base model for manipulation), we choose the model within the Rashomon set that has full row rank over distinct leaf patterns in the training and validation sets and has the lowest cross-validated log loss. We define a group to be a set of training and validation rows that land in the same leaf in every tree, meaning they always change together. On Adult, each rank check is costly, so we accept a model with rank 39,028 of 39,029. See Table~\ref{tab:hard-start} for starting model details.

\begin{table}[t]
\centering
\setcounter{table}{0}
\renewcommand{\thetable}{S\arabic{table}}
\caption{Starting models for hard manipulation. A group is a set of training and validation rows that fall in the same leaf of every tree. $\Phi$ is the $n \times m$ 0/1 matrix with $\Phi_{j,k}=1$ if group $j$ lands in leaf $k$.}
\label{tab:hard-start}
\begin{tabular}{lrrrr}
\toprule
 & Trees $T$ & Leaves $m$ & Groups $n$ & Rank of $\Phi$ \\
\midrule
HELOC  & 283    & 12{,}748  & 7{,}545  & 7{,}545  \\
COMPAS & 24{,}000 & 616{,}100 & 2{,}164  & 2{,}164  \\
Adult  & 1{,}645  & 208{,}915 & 39{,}029 & 39{,}028 \\
German & 244    & 19{,}061  & 800      & 800      \\
\bottomrule
\end{tabular}
\end{table}

Every individual lands in one leaf per tree, and the final prediction is a logistic transformation of the sum of these leaf values. To make the manipulation, we find the smallest total change (minimum L2 norm) that shifts a single row in the validation set without changing any other group's prediction or taking the model outside of the Rashomon set. We repeat this for each row in the validation set. Formally, we define the starting model's slack to be $\sigma_B = (1+\varepsilon)B_{ref} - B_0$ where $B_{ref}$ is the validation Brier of the reference model and $B_0$ is the starting model's Brier score. We send the target value across the threshold ($t = 0.5$) towards the preset cap of $0.9$ (if $p < 0.5$) or $0.1$ (if $p > 0.5$), or stop early if the Brier score changes more than $0.999\sigma_B$. Note that hard manipulation can yield implausibly large model files with high numbers of leaves or trees---this is in part why we say hard manipulation can be detected given access to code, data, and/or the model file.

\subsection{Soft manipulation: multi-metric evaluation}
\label{supp:multi-metric}
We added area under the ROC curve (AUC) as a check in multi-metric evaluation, where a model passes if its $1-\text{AUC}$ is within a tolerance of $\varepsilon$ of the best $1 - \text{AUC}$. AUC is the probability that our model will correctly predict a higher probability for a randomly chosen positive case than a randomly chosen negative case. Calibration-in-the-large is calculated as $|\frac{\text{Mean Predicted Probability}}{\text{Mean Observed Outcome}} - 1| = |\frac{\frac{1}{n}\sum_{i = 1}^{n} \hat{p_i}}{\frac{1}{n}\sum_{i = 1}^{n} Y_i} - 1|$, and passes the check if it is within $0.1$ of the smallest calibration-in-the-large value. For calibration slope, we use $|\text{slope} - 1| = |b - 1|$ and calibration slope within $0.1$ of the lowest $|b-1|$ as our check, where slope $b$ is the coefficient from the following model: $P(y=1) = \text{sigmoid}(a + b \cdot \text{logit}(\hat{p}))$. $b$ is a measure of the model's confidence; $b < 1$ means that the model's predictions are close to 0 and 1 (too extreme), indicating model overconfidence, and $b > 1$ means that the model's predictions are too bunched around 0.5, indicating underconfidence. A candidate model fails the impact ratio check if the ratio is significantly worse than the reference model's, which is defined using 1000 paired bootstrap resamples of the validation rows and a one-sided test of mean difference in ratio ($\alpha$ = 5\%). Subgroups are sex on COMPAS, Adult, and German datasets, with additional Black and White subgroups on the COMPAS and Adult (includes other) datasets. Groups with fewer than 30 validation rows are omitted. There are no subgroups on the HELOC dataset, so each model only needs to meet the other criteria to be considered within the Rashomon set. Favorable decisions are no re-arrest for COMPAS, income above \$50,000 on Adult, and good credit on German.

For the wildfire data, we use a more relaxed AUC check than for the other datasets, where a model passes if its validation AUC is within 0.01 of the best AUC. Because wildfire AUCs are high (the best is 0.894), the prior rule of $1-\text{AUC}$ within $\varepsilon$ of the best model's $1-\text{AUC}$ allows a gap of only about 0.001, and excludes every wildfire model, including the reference model (AUC 0.892). Because the wildfire data have no protected attribute, we do not conduct an analysis for impact ratio.

\subsection{Wildfire: hard manipulation}
\label{supp:wildfire-hard}
In the aggregate estimate case, we aim to shift the mean of the predicted probabilities in the pre-industrial scenario. Here, we do not hold present-day predictions fixed. Edits are limited by whether they remain within the Rashomon set. Specifically, we require the cross-validated Brier score to stay within the 1\% tolerance, as well as each fold model's Brier score on its own training years to stay within the 1\% tolerance. This is a conservative approach that restricts both the training loss and cross-validation loss---we do this so the edit cannot be trivially cherry-picked by fitting the held-out years' outcomes. In practice, a motivated actor could do so with no loss in reported predictive performance. The edits are found numerically, via penalizing edits that break the tolerance limits. We check the results by rebuilding the models with the edited leaf values.

\subsection{Reconcile: procedure}
\label{supp:reconcile}
Each model in the Rashomon set is reconciled with the reference model. Using validation set B, Reconcile finds the rows where the prediction from the reference model is $\tau = 0.05$ greater than the prediction from the comparison model and vice versa. For each of the two regions, it considers a correction to either the reference model's predictions or the comparison model's predictions using the true outcome rate on validation set B. It chooses the correction which decreases the Brier score by the most, which is a function of the share of rows that are being shifted and the size of the shift. It additionally shifts any predictions in the validation set A or test set that meet the criteria for shifting: differing by $\ge \tau$ in the same direction. This continues until at most 2\% ($\alpha = 0.02$) of the predictions in validation set B differ by more than $\tau$, or after 500 rounds, or no patch improves the Brier score. 

Note that when we compare the adversarially manipulated reconciliations against ones at random, we do not condition on disagreement. Disagreement is a gain arising from inducing adversarial multiplicity, and conditioning on it would attenuate the estimated effect of manipulation (e.g., if most models in the Rashomon set do not disagree with the reference model, then conditioning only on models that disagree would not capture the impact the adversary has by finding a disagreeing model in the first place).

\section{Policies and rulings behind Figure~\ref{fig:policy}}
  \providecommand{\policyoverviewcaption}{Policies and court rulings behind Figure~\ref{fig:policy}. ``Disclosure level'' describes what an auditor, regulator, court, affected individual or the public is guaranteed, and what it can obtain only on request or in some cases.}
{\footnotesize\setlength{\tabcolsep}{3pt}\renewcommand{\arraystretch}{1.1}
\begin{longtable}{@{}>{\raggedright\arraybackslash}p{0.3\textwidth}>{\raggedright\arraybackslash}p{0.15\textwidth}>{\raggedright\arraybackslash}p{0.5\textwidth}@{}}
\caption{\policyoverviewcaption}\label{tab:policyoverview}\\
\toprule
\textbf{Policy} & \textbf{Jurisdiction} & \textbf{Disclosure level} \\
\midrule
\endfirsthead
\toprule
\textbf{Policy} & \textbf{Jurisdiction} & \textbf{Disclosure level} \\
\midrule
\endhead
\bottomrule
\endfoot
\multicolumn{3}{@{}l}{\textbf{Insurance}} \\*
EU AI Act (high-risk) & EU & Documentation; the model and its training data on request or in some cases \\
GDPR Arts. 15, 22, 58 & EU & Documentation; the model and its training data on request or in some cases \\
UK DUAA s.80 & UK & Documentation; the model and its training data on request or in some cases \\
NAIC AI bulletin & US states & Documentation \\
NAIC CASTF white paper & US states & Documentation; the model or its formula on request or in some cases \\
Colorado SB21-169 & Colorado & Documentation; black box testing on request or in some cases \\
Colo. Reg. 10-1-1 & Colorado & Documentation \\
NY DFS CL 2024-7 & New York & Documentation; black box testing on request or in some cases \\
Cal. Prop 103 rates & California & Documentation \\
Cal. auto class plans & California & The model or its formula; the model and its training data on request or in some cases \\
\emph{Lokken v. UnitedHealth} (ruling) & US federal & Access in the case: documentation \\
\addlinespace[4pt]
\multicolumn{3}{@{}l}{\textbf{Forensic DNA analysis}} \\*
SWGDAM PG 2015 & US federal & Organization's own review only \\
FBI QAS (DNA labs) & US federal & Documentation \\
ANSI/ASB Std 018 & US & Organization's own review only \\
\emph{State v. Pickett} (ruling) & New Jersey & Access in the case: source code \\
\emph{U.S. v. Ellis} (ruling) & US federal & Access in the case: source code \\
\emph{Virginia v. Watson} (ruling) & Virginia & Access in the case: source code \\
\emph{U.S. v. Johnson (FST)} (ruling) & US federal & Access in the case: source code \\
\emph{U.S. v. Gissantaner} (ruling) & US federal & Access in the case: source code \\
\emph{U.S. v. Anderson} (ruling) & US federal & Access in the case: black box testing \\
\emph{People v. Wakefield} (ruling) & New York & Access in the case: documentation \\
\emph{People v. Chubbs} (ruling) & California & Access in the case: documentation \\
\addlinespace[4pt]
\multicolumn{3}{@{}l}{\textbf{Public sector AI}} \\*
OMB M-25-22 & US federal & Agency's own review only \\
OMB M-25-21 & US federal & Agency's own review only \\
EO 13960 & US federal & Agency's own review only \\
France CRPA & France & Documentation; source code on request or in some cases \\
Canada DADM & Canada & Documentation; black box testing on request or in some cases \\
EU AI Act (high-risk) & EU & Documentation; the model and its training data on request or in some cases \\
GDPR Arts. 15, 22, 58 & EU & Documentation; the model and its training data on request or in some cases \\
UK DUAA s.80 & UK & Documentation; the model and its training data on request or in some cases \\
\emph{Houston Fed. Teachers} (ruling) & US federal & Access in the case: documentation \\
\addlinespace[4pt]
\multicolumn{3}{@{}l}{\textbf{Hiring tools}} \\*
Uniform Guidelines & US federal & Documentation; the model and its training data on request or in some cases \\
CT SB 5 & Connecticut & Documentation \\
IL HB 3773 & Illinois & Documentation \\
CA CRC ADS regs & California & Documentation \\
NYC LL 144 & New York City & Documentation; black box testing on request or in some cases \\
EU AI Act (high-risk) & EU & Documentation; the model and its training data on request or in some cases \\
GDPR Arts. 15, 22, 58 & EU & Documentation; the model and its training data on request or in some cases \\
UK DUAA s.80 & UK & Documentation; the model and its training data on request or in some cases \\
\addlinespace[4pt]
\multicolumn{3}{@{}l}{\textbf{Credit scoring}} \\*
ECOA / Reg. B & US federal & Documentation \\
FCRA key factors & US federal & Documentation \\
EU AI Act (high-risk) & EU & Documentation; the model and its training data on request or in some cases \\
AI Act Art. 86 & EU & Documentation \\
GDPR Arts. 15, 22, 58 & EU & Documentation; the model and its training data on request or in some cases \\
EU Consumer Credit Dir & EU & Documentation \\
UK DUAA s.80 & UK & Documentation; the model and its training data on request or in some cases \\
\addlinespace[4pt]
\multicolumn{3}{@{}l}{\textbf{Expert evidence in court}} \\*
FRCP 26 (experts) & US federal & Documentation; code and data to rerun training on request or in some cases \\
FRE 702 (2023) & US federal & Documentation \\
FRE 706 & US federal & Documentation \\
Fed. R. Crim. P. 16 & US federal & Documentation; code and data to rerun training on request or in some cases \\
Cal. Evid. 1060-62 & California & Documentation; source code on request or in some cases \\
N.D. Cal. model PO & US federal & Documentation; source code on request or in some cases \\
\addlinespace[4pt]
\multicolumn{3}{@{}l}{\textbf{ADMT laws}} \\*
TX TRAIGA & Texas & Documentation \\
Colorado SB 26-189 & Colorado & Documentation \\
CCPA ADMT rules & California & Documentation; black box testing on request or in some cases \\
GDPR Arts. 15, 22, 58 & EU & Documentation; the model and its training data on request or in some cases \\
UK DUAA s.80 & UK & Documentation; the model and its training data on request or in some cases \\
\addlinespace[4pt]
\multicolumn{3}{@{}l}{\textbf{Bank models}} \\*
SR 26-2 & US federal & Organization's own review only \\
EU CRR (IRB) & EU & Documentation \\
EU IRB RTS 2022/439 & EU & Documentation; code and data to rerun training on request or in some cases \\
UK PRA SS1/23 & UK & Organization's own review only \\
\addlinespace[4pt]
\multicolumn{3}{@{}l}{\textbf{Bail and sentencing risk tools}} \\*
Idaho \S{}19-1910 & Idaho & The model and its training data \\
Cal. Penal 1320.35 & California & The model or its formula \\
EU AI Act (high-risk) & EU & Documentation; the model and its training data on request or in some cases \\
\emph{State v. Loomis} (ruling) & Wisconsin & Access in the case: documentation \\
\addlinespace[4pt]
\multicolumn{3}{@{}l}{\textbf{Peer-reviewed research}} \\*
Journal practice & Varies & Documentation; a pre-registered analysis on request or in some cases \\
\end{longtable}}

\end{document}